\documentclass[10pt,journal,compsoc]{IEEEtran}
\ifCLASSOPTIONcompsoc
  \usepackage[nocompress]{cite}
\else
  \usepackage{cite}
\fi
\usepackage{booktabs,subcaption,amsfonts,dcolumn}
\usepackage{graphicx}
\usepackage{float}
\usepackage{pifont}
\usepackage{amsmath}
\usepackage{amssymb}

\usepackage[colorlinks,bookmarksopen,bookmarksnumbered,citecolor=blue, linkcolor=blue, urlcolor=blue]{hyperref}
\usepackage{graphicx}
\usepackage{media9}
\usepackage{xcolor}   
\usepackage{bbding}   
\usepackage{gensymb}
\usepackage{mathtools}
\usepackage{multirow}
\usepackage{booktabs}
\usepackage{amsthm}
\usepackage{braket}
\ifCLASSINFOpdf
\else
\fi
\begin{document}
%
\title{VALOR: Vision-Audio-Language Omni-Perception Pretraining Model and   Dataset}
%
%
%
%


\author{ Jing~Liu*, Sihan~Chen*, Xingjian~He, Longteng~Guo, Xinxin~Zhu, Weining~Wang, \\ Jinhui~Tang, ~\IEEEmembership{Senior Member,~IEEE}  


\IEEEcompsocitemizethanks{\IEEEcompsocthanksitem Jing Liu (corresponding author) and Sihan Chen contribute equally to this paper, and they are  with School of Artificial Intelligence, University of Chinese Academy of Sciences and  National Laboratory of Pattern Recognition, Institute of Automation, Chinese Academy of Sciences. Email: \{jliu, sihan.chen\}@nlpr.ia.ac.cn. Xingjian He, Longteng Guo, XinXin Zhu and Weining Wang are with  National Laboratory of Pattern Recognition, Institute of Automation, Chinese Academy of Sciences. Email: longtengguo1993@gmail.com, \{xingjian.he, xinxin.zhu, weining.wang\}@nlpr.ia.ac.cn. Jinhui Tang is with Nanjing University of Science and Technology, School of Computer Science and Engineering. Email: jinhuitang@njust.edu.cn}}

\markboth{Journal of \LaTeX\ Class Files,~Vol.~14, No.~8, August~2015}%
{Shell \MakeLowercase{\textit{et al.}}: Bare Demo of IEEEtran.cls for Computer Society Journals}
%



\IEEEtitleabstractindextext{%
\begin{abstract}

In this paper, we propose a \textbf{V}ision-\textbf{A}udio-\textbf{L}anguage \textbf{O}mni-pe\textbf{R}ception pretraining model  (VALOR) for multi-modal understanding and generation. Different from widely-studied vision-language pretraining models, VALOR jointly models relationships of vision, audio and language  in an end-to-end manner.  It contains three separate encoders for  single modality representations, and a  decoder for multimodal conditional text generation. We design two pretext tasks to pretrain VALOR model, including Multimodal Grouping Alignment  (MGA) and Multimodal Grouping Captioning (MGC). MGA projects vision, language and audio to the same common space,  building vision-language, audio-language and  audiovisual-language alignment simultaneously. MGC learns how to generate text tokens in conditions of  vision, audio or their both. To promote vision-audio-language pretraining research, we construct a large-scale high-quality tri-modality dataset  named VALOR-1M, which contains 1M audiable videos  with human annotated  audiovisual captions. Extensive experiments show that VALOR
can learn strong multimodal correlations  and  be generalized to various   downstream tasks  (e.g., retrieval, captioning and question answering), with different input modalities  (e.g., vision-language, audio-language and audiovisual-language). VALOR achieves new state-of-the-art performances  on series of public cross-modality benchmarks. Code and data are available at project page  \url{https://casia-iva-group.github.io/projects/VALOR}.

\end{abstract}

}

\maketitle

\IEEEdisplaynontitleabstractindextext

%
\IEEEpeerreviewmaketitle

\IEEEraisesectionheading{\section{Introduction}\label{sec:introduction}}

 As human beings, we perceive information from  environment through  multiple mediums  (e.g. looking, reading, hearing, touching or smelling), and  further understand or interact with the world  based on those multimodal clues.  An ideal intelligent system should also imitate this, to develop both cross-modal understanding and generation capabilities.  Various cross-modality applications has been extensively studied, among which vision-language tasks take the main part, including text-to-vision retrieval\cite{kim2018bilinear,dong2021dual}, vision captioning\cite{liu2021aligning, yan2019fine,wang2019controllable} and visual question answering\cite{anderson2018bottom,peng2020mra}.  Fortunately, inspired by  the great success of self-supervised pretraining methods in natural language processing\cite{devlin2018bert,radford2018improving,brown2020language},  vision-language pretraining  has developed rapidly, and achieved dominated performances  over traditional methods on various of vision-language benchmarks.

\begin{figure}[t]
\centering
\includegraphics[width=1.0\linewidth]{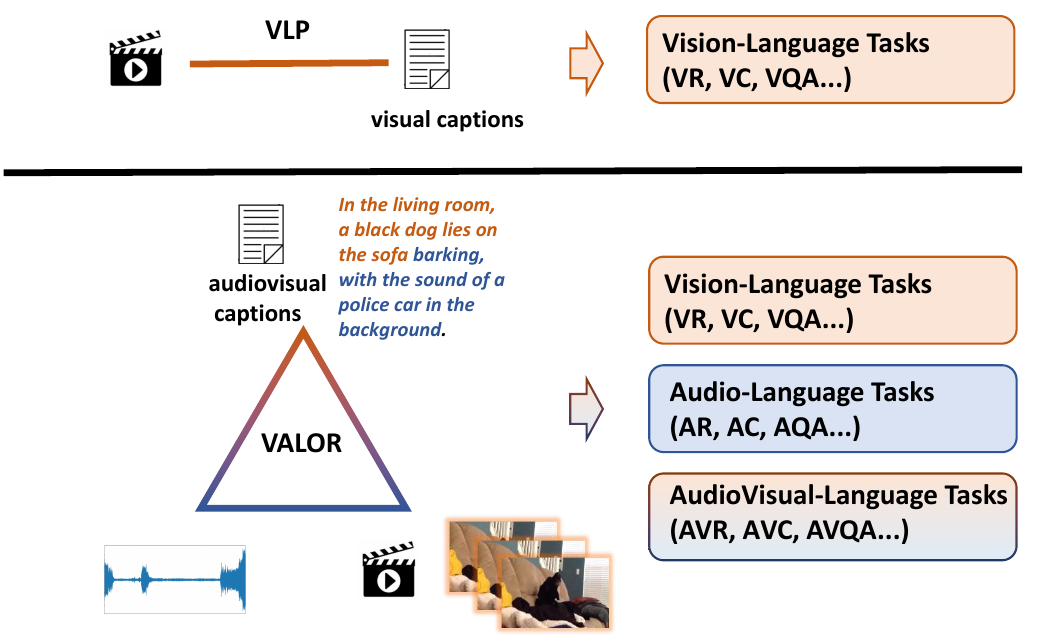}
\caption{VALOR takes correlated vision-audio-language  data for pretraining, and can generalize to multiple tasks. AVR/VR/AR represent text-to-audiovisual/visual/audio retrieval, AVC/VC/AC represent audiovisual/visual/audio captioning, and AVQA/VQA/AQA represent audiovisual/visual/audio question answering, respectively. Click the botton to play the audio.
}
\label{fig:zongti}
\end{figure}

 However, we argue that modeling  relationship between vision and language is far from enough  to establish a powerful multimodal system and additionally introducing audio modality  to build tri-modality interactions is necessary. On one hand, audio signal usually contains semantic meanings complementary to vision, and thus utilizing three modalities  can help machine understand aroundings more comprehensively and accurately. As the example in Figure \ref{fig:zongti} shown,   we can only know what's going on inside the room through observing video frames,  but miss perceptions about the outside police car  unless we  hear the police siren. On the other hand, modeling three modalities in a unified end-to-end framework can  enhance model's generalization capabilities, and benefit various of vision-language, audio-language, audiovisual-language and vision-audio downstream tasks.
 
 \begin{figure*}[t]
\centering
\includegraphics[width=0.6\linewidth]{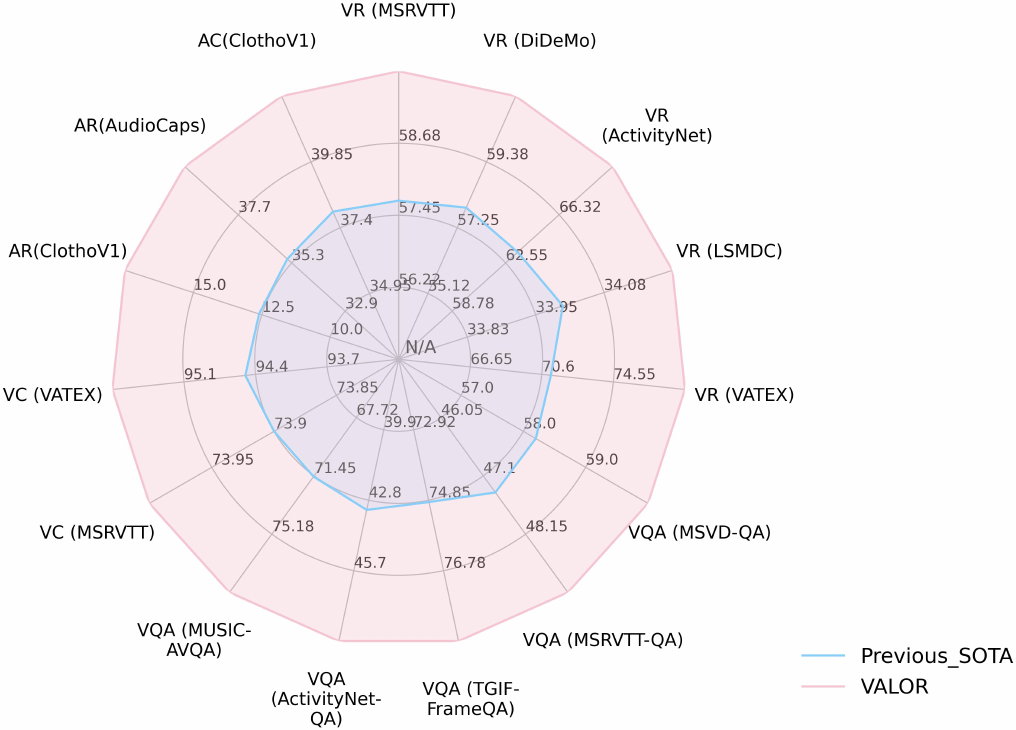}
\caption{VALOR achieves state-of-the-art performances on a broad range of tasks compared with
other customized or foundation models. VR, VC, VQA represent text-to-video retrieval, video captioning and video QA, respectively.}

\label{fig:rador}
\end{figure*}
 To this end, as shown in Figure \ref{fig:zongti}, we propose a \textbf{V}ision-\textbf{A}udio-\textbf{L}anguage \textbf{O}mni-pe\textbf{R}ception pretraining model  (VALOR) to build universal connections among  three modalities, and to fulfill tri-modality understanding and generation.  As shown in Figure \ref{fig:model_architecture}, VALOR encodes vision, audio  and language separately with three single-modality encoders, and use  a multimodal decoder for conditional text generation.  Two pretext tasks, i.e., Multimodal Grouping Alignment  (MGA) and Multimodal Grouping Captioning  (MGC) are designed to endow VALOR  with the capabilities to tackle both discriminative and generative tasks.  Specifically, MGA  projects three modalities into the same common space, and establishes fine-grained alignment between three modality groups including vision-language, audio-language and audiovisual-language  via contrastive learning. MGC demands models to reconstruct randomly masked text tokens, conditioned by vision, audio, or their both via cross attention layers. Thanks to modality grouping strategy, VALOR can learn how to align  or generate text  according to different modality combinations, and such capabilities can be transferred to various kinds of cross-modality downstream tasks, including video/audio/audiovisual retrieval, captioning or question answering.

\begin{figure*}[ht]
\centering
\includegraphics[width=1.0\linewidth]{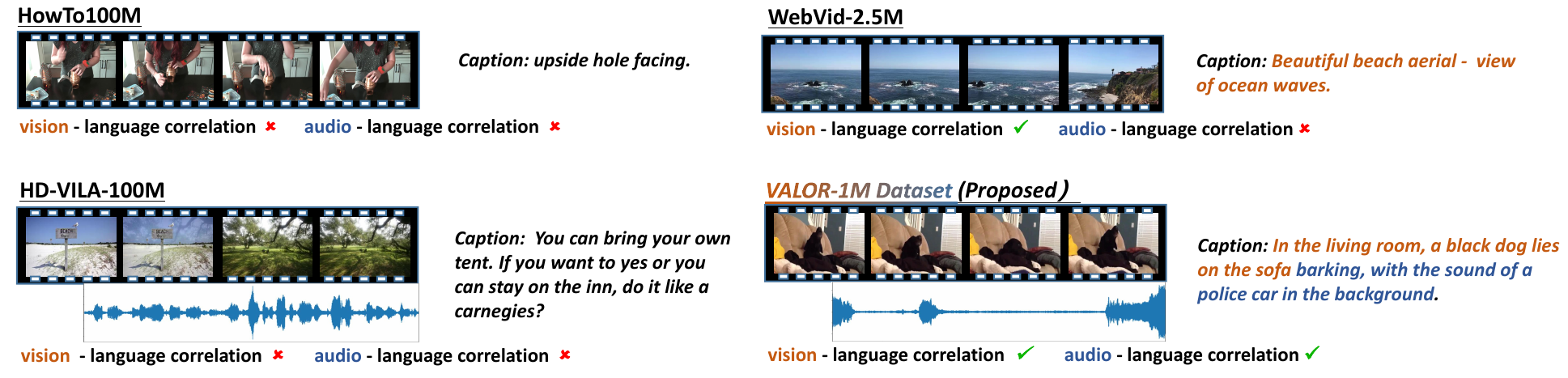}
\caption{Visualizations of  video-language pretraining datasets including HowTo100M\cite{miech2019howto100m}, HD\_VILA\_100M\cite{xue2022advancing}, WebVid-2,5M\cite{bain2021frozen} and VALOR-1M. Click the bottons to play the audio.}
\label{fig:data representaion}
\end{figure*}

In addition, we argue that strong correlated vision-audio-language  triplets are indispensable for training strong tri-modality models. Current public vision-language datasets are incapable of tri-modality pretraining for that \romannumeral1) all image-language datasets and some video-language datasets like WebVid-2.5M\cite{bain2021frozen} do not contain audio signals. \romannumeral2) Even if some video-language datasets like HowTo100M\cite{miech2019howto100m} and HD\_VILA\_100M\cite{xue2022advancing} contain audio modality,  their audios are limited to human speech with less diversity, and their texts are ASR transcriptions instead of objective descriptions, which are overlapped with speech. To overcome above restrictions,  we construct a large-scale high-quality vision-audio-language dataset (VALOR-1M) to promote tri-modality pretraining researches. It contains one million open-domain audiable videos, each of which is manually annotated with one  audiovisual caption,  describing  both audio and visual contents simultaneously. VALOR-1M's  strong vision-language and audio-language correlations, and its large scaling make it the best choice for tri-modality  pretraining. Besides VALOR-1M, we also establish a new benchmark VALOR-32K for evaluations on audiovisual-language capabilities. It  contains two new tasks, including audiovisual-retrieval (AVR) and audiovisual captioning (AVC).

Extensive ablation studies have been conducted to  demonstrate  effectiveness of proposed VALOR model and modality grouping strategy. Both quantitative  and qualitative results prove that VALOR can utilize audiovisual clues for AVR and AVC tasks effectively. We  extensively validate VALOR on series of public video-language, image-language and audio-language benchmarks, and it achieved series of  new state-of-the-art results. Specifically, as shown in Figure \ref{fig:rador}, VALOR outperforms previous state-of-the-art methods by 3.8\%, 6.2\%, 12.7\%, 0.6\%,  10.4\% (R@1) on  text-to-video retrieval benchmarks including MSRVTT, DiDeMo, ActivityNet, LSMDC and VATEX; 3.8\%, 3.4\%, 5.1\%, 12.5\% (Acc)  on Open-ended video question answering benchmarks including MSRVTT-QA, MSVD-QA, TGIF-FrameQA and ActivityNet-QA; 38.9\%, 13.0\% (R@1) on text-to-audio retrieval benchmarks including ClothoV1 and AudioCaps. In addition, VALOR outperforms GIT2 big model\cite{wang2022git} on VATEX captioning benchmark with only 0.26\% training data and 11.6\% parameters.

Overall, the contribution of this work can be summaried as follows:

 \uppercase\expandafter{\romannumeral1}) We proposed an omni-perception pretraining model  (VALOR), which establishes correlations among  vision, audio and  language for tri-modality understanding and generation. 
 
 \uppercase\expandafter{\romannumeral2}) We introduced  MGA and MGC pretraining tasks with modality grouping strategy to enhance model's  generalization capability with  different modality inputs.

 \uppercase\expandafter{\romannumeral3}) We proposed VALOR-1M dataset which is  the first large-scale human-annotated  tri-modality dataset to promote vision-audio-language  researches, and VALOR-32K benchmark for evaluations on audiovisual-language capabilities.

 \uppercase\expandafter{\romannumeral4}) Pretrained on VALOR-1M and current public vision-language datasets, VALOR has achieved new state-of-the-art performances on series of cross-modality benchmarks with evident improvements.

\section{Related Work}
In this section, we first introduce common cross-modality datasets used for multimodal pretraining. After that we review vision-language  pretraining methods. At last, we introduce typical methods utilizing more modalities beyond vision and text for  video-language learning.

\subsection{Cross-Modality Datasets for Multimodal Pretraining}
Generally, an ideal vision-language pretraining dataset  should meets two basic demands, large scaling enough and strong visual-textual correlations. Considering that  sentence-level caption annotations are much more resource-consuming than word-level label tagging, some  methods attempts to collect videos which contains human speech, and extract ASR transcriptions  as captions. For example,   Miech et al. collected 
 HowTo100M\cite{miech2019howto100m}, which consists of 136M video clips sourced from 1.22M narrated instructional YouTube videos, and it has  become the main-stream dataset used by early video-language pretraining methods.  Zellers et al. followed this approach and proposed   YT-Temporal-180M \cite{zellers2022merlot} which contains 180M clips from 6M YouTube videos.  Xue et al. collected HD\_VILA\_100M\cite{xue2022advancing} that consists of 100M clips from 3.3M YouTube videos, with  more diversity and larger image resolution. 

However, although this route can be friendly scaled up to get large amount of  video-text pairs, the quality of captions are  not satisfying. Besides probable speech recognition errors,  ASR transcriptions usually convey subjective ideas and opinions  of speechers, instead of  objective descriptions of static objects and happening events.  Even if some  transcriptions  indeed reflect visual contents, there  exists  temporal misalignment problem that they may correspond to video clips before or after\cite{miech2020end}. To overcome this problem and  pursue both quantity and quality, Bain et al. followed  collection procedures of image-language  Conceptual Captions datasets (CC3M\cite{sharma2018conceptual}, CC12M\cite{changpinyo2021conceptual}), and collected  WebVid\cite{bain2021frozen}, which consists of 2.5M videos paired with alt-texts. Although sometimes unfluent and incomplete, alt-texts have overall stronger correlations to video contents than ASR transcriptions, and have been widely used by latest video-language pretraining methods. However, none of datasets mentioned above support vision-audio-language pretraining, due to  the missing of audio-language correlations, which motivates us to collect VALOR-1M dataset to push tri-modality  pretraining development.

 \subsection{Vision-Language Pretraining}

Influenced by the success of BERT\cite{devlin2018bert}, vision-language pretraining has got rapid development, we summarize serveral main research directions as following. 

 \uppercase\expandafter{\romannumeral1}) \textbf{Cross-Modality Pretraining Framework Design.} According to different network architectures, vision-language models can be mainly divided into dual-encoder paradigm\cite{seo2022end, xue2022clip} and fusion-encoder paradigm\cite{chen2020uniter,lu2019vilbert}. The former fuses vision and language lightly at the output of encoders by simple dot-product, which can be efficiently used for cross modality retrieval and zero-shot classification. The Latter use co-attention\cite{lu2019vilbert} or merge-attention\cite{chen2020uniter} to fuse two modalities deeply, which are good at more fine-grained tasks like captioning or VQA. In addition,  various of self-supervised pretext tasks have been proposed for better cross-modality feature representation learning, including masked language modeling (MLM)\cite{chen2020uniter}, masked vision modeling (MVM)\cite{chen2020uniter,fu2022empirical}, vision-text matching (VTM)\cite{chen2020uniter,li2020hero}, vision-text contrastive learning (VTC)\cite{li2021align,bain2021frozen}, etc. With regards to visual representations, early methods separately use off-line object detectors  (e.g., Faster-RCNN\cite{ren2015faster}) to extract object-level image features or 3D convolutional neural networks  (e.g.,  S3D\cite{seo2022end}) to extract clip-level video features.  With the emerging of vision transformers\cite{dosovitskiy2020image,liu2021swin}, image-language and video-language can be unified by  feeding models images or sparsely sampled frames. 

\uppercase\expandafter{\romannumeral2}) \textbf{Unified Multi-Task Modeling.}  This series of works attempts to universally model different tasks with a unified framework and remove task-specific finetuning heads, to  utilize pretraining data more efficiently. VL-T5\cite{cho2021unifying} first uses a sequence-to-sequence framework to model vision-language tasks like VQA and viusal grounding. Later, fine-grained localization tasks like object detection and text-to-image generation are also integrated\cite{yang2022unitab,wang2022ofa,lu2022unified}, through box coordinates tokenization\cite{Chen2021Pix2seqAL} or image tokenization\cite{van2017neural}, respectively. Besides sequence-to-sequence framework, some works also unify multiple vision-language tasks via contrastive learning\cite{zhu2022uni} or masked language modeling\cite{li2022lavender}. However, even if above methods have unified multiple tasks, they are constrained in vision-language domain. In comparison, VALOR can generalize to vision-audio-language domain, and suitable for partial- and omni- perception tasks.

\uppercase\expandafter{\romannumeral3}) \textbf{Vision-Language Foundation Models.} Vision-language models trained with extremely huge  data  and parameters are usually called big models or foundation models, and are often supervised with contrastive learning\cite{radford2021learning,jia2021scaling,yuan2021florence,pham2021combined}, language modeling\cite{wang2021simvlm,alayrac2022flamingo,wang2022git,chen2022pali} , or both\cite{yu2022coca}. Foundation models have achieved dominated performances on vision-language benchmarks. For example, Flamingo\cite{alayrac2022flamingo} increases  model size to 80.2B parameters and got 84.0 Acc score on VQAv2 dataset, while  GIT2\cite{wang2022git}  increases  data size to 12.9B image-text pairs and achieved 149.8 CIDEr score on COCO caption benchmark. However,  due to  high demands on computing resources, data storage and complicated distributed training, scaling vision-language pretraining models from parameter and data dimensions shows limited efficiency. In comparison, we assume that VALOR can be viewed as scaling up from  modality dimension, by introducing audio  and building tri-modality connections, which is effective and more efficient.

\subsection{Auxiliary Modality Enhanced Video-Language Understanding}

Considering videos are naturally multimodal medium and each modality contains rich semantic meanings, some approaches exploited  more modalities to enhance video-language learning.  MMT\cite{gabeur2020multi} proposes a multimodal transformer to fuse seven modality experts for text-to-video retrieval. SMPFF\cite{chen2021mm21} additionally introduce objective and audio features to improve video captioning. In  large-scale pretraining scenario, audio and subtitle are the most commonly used  auxiliary modalities to strengthen video representation. UniVL\cite{luo2020univl}, VLM\cite{xu2021vlm} and MV-GPT\cite{seo2022end} fuse video and subtitle modalities, and pretrain on HowTo100M dataset for video captioning. VALUE\cite{li2021value} further exploit  subtitle enhancement on more tasks including video retrieval and QA. With regards to audio enhancement, AVLNet\cite{rouditchenko2020avlnet} and  MCN\cite{chen2021multimodal} utilize audio to enhance text-to-video retrieval. VATT\cite{akbari2021vatt}  proposed a hierarchical contrastive loss for text-video and video-audio alignment, but it targets at learning single-modality representations instead of improving cross-modality capabilities.  MERLOT Reserve\cite{zellers2022merlot} and i-Code\cite{yang2022code} also take vision, audio and language  as input for pretraining, but has essential differences with VALOR in that  \romannumeral1 ) those methods has severe pretraining-finetuning inconsistency. Specifically, the audio-language relation  are between human speech and ASR transcriptions during pretraining, but general audios and objective descriptions during finetuning.  By contrast, VALOR is trained on strong correlated tri-modality dataset and keeps pretraining-finetuning consistency, which makes it can generalize to video-language, audio-language and audiovisual-language tasks. \romannumeral2 ) those methods only targets at discriminative tasks like video QA, while VALOR can tackle discriminative, contrastive and generative tasks, thanks to the unified architecture and designed pretraining tasks.

\begin{table*}[]
\end{table*}

\begin{table*}
\centering
\caption{Statistics of common public video-language pretraining datasets  and downstream benchmark datasets.   Audio: dataset contains audio or not.  V-L: vision-language correlation. A-L: audio-language correlation. \#Example: the number of videos/audios/images. \#Clips: the number of video clips or audio clips. $\rm Len_{Cap}$: average caption length. ACD: audio concepts density.}
\label{data statistics}
\scalebox{0.9}{
\begin{tabular}{lllllllllll}
\toprule
Dataset               & Caption   & Task       &  Domain  &  Audio                  &  V-L                 &  A-L                 & \#Example & \#Clips  & $\rm Len_{Cap}$ & ACD (\%) \\
\midrule
\multicolumn{4}{l}{\textbf{\textit{Pretraining Datasets}}} \\
HowTo100M\cite{miech2019howto100m}              & ASR       & -           & Instructional & \Checkmark  & \ding{55} & \ding{55} & 1.22M    & 136M           & 4.0        & 3.4     \\
HD\_VILA\_100M\cite{xue2022advancing}          & ASR       & -           & Open          & \Checkmark  & \ding{55} & \ding{55} & 3.3M     & 103M           & 32.5       & 1.1     \\
WebVid-2.5M\cite{bain2021frozen}              & Alt-text & -           & Open          & \ding{55} & \Checkmark  & \ding{55} & 2.5M     & 2.5M             & 14.2       & 3.3     \\
CC3M\cite{sharma2018conceptual} & Alt-text & -           & Open          & \ding{55} & \Checkmark  & \ding{55} & 3.3M        & -       & -                & 3.0     \\
\textbf{VALOR-1M}              & Manual    & -           & Open          & \Checkmark  & \Checkmark  & \Checkmark  & 1.18M    & 1.18M            & 16.4       & 9.7     \\

\midrule
\multicolumn{4}{l}{\textbf{\textit{Downstream Benchmarks}}} \\
MSVD\cite{chen2011collecting}                    & Manual    & VR,VC,VQA   & Open          & \ding{55} & \Checkmark  & \ding{55} & 2K       & 2K                & 7.0        & 6.7     \\
MSRVTT\cite{xu2016msr}                 & Manual    & VR,VC,VQA & Open          & \Checkmark  & \Checkmark  & \ding{55} & 7K       & 10K              & 9.3        & 4.7     \\
VATEX\cite{wang2019vatex}                  & Manual    & VR,VC       & Open          & \Checkmark  & \Checkmark  & \ding{55} & 41.3K    & 41.3K            & 14.3       & 4.1     \\
YouCook2\cite{zhou2018towards}               & Manual    & VR,VC       & Cooking       & \Checkmark  & \Checkmark  & \ding{55} & 2K       & 15.4K         & 8.8        & 4.3     \\
DiDeMo\cite{anne2017localizing}                 & Manual    & VR          & Open          & \Checkmark  & \Checkmark  & \ding{55} &   10.5K       & 26.9K               & 8.0        & 6.3     \\
ActivityNet\cite{krishna2017dense}            & Manual    & VR,VC       & Action        & \Checkmark  & \Checkmark  & \ding{55} & 20K      & 100k           & 13.5       & 3.4     \\
LSMDC\cite{rohrbach2017movie}                  & Manual    & VR          & Movie         & \Checkmark  & \Checkmark  & \ding{55} &      202    & 118K           & 9.1        & 2.5     \\
ClothoV1\cite{drossos2020clotho}                  & Manual    & AR,AC          & Open         & \Checkmark  &  \ding{55}  & \Checkmark&      5.0K    & 5.0K           & 11.3        & 10.4     \\
 
AudioCaps\cite{kim2019audiocaps}                  & Manual    & AR,AC          & Open         & \Checkmark  &  \ding{55}  & \Checkmark&      51.3K    & 51.3K           & 8.8        & 17.3     \\
 
Pano-AVQA\cite{yun2021pano}              & Manual    & AVQA        & Panoramic     & \Checkmark  & \Checkmark  & \Checkmark  & 5.4K     & 5.4K            & -          & -       \\
MUSIC-AVQA\cite{li2022learning}             & Manual    & AVQA        & Music         & \Checkmark  & \Checkmark  & \Checkmark  & 9.3K     & 9.3K           & -          & -       \\
AVQA\cite{yang2022avqa}                    & Manual    & AVQA        & Open          & \Checkmark  & \Checkmark  & \Checkmark  & 57K      & 57K             & -          & -       \\
\textbf{VALOR-32K}             & Manual    & AVR, AVC     & Open          & \Checkmark  & \Checkmark  & \Checkmark  & 32K      & 32K             & 19.8       & 9.1    \\
\bottomrule
\end{tabular}}
\end{table*}

\begin{figure}[ht]
\centering
\includegraphics[width=1.0\linewidth]{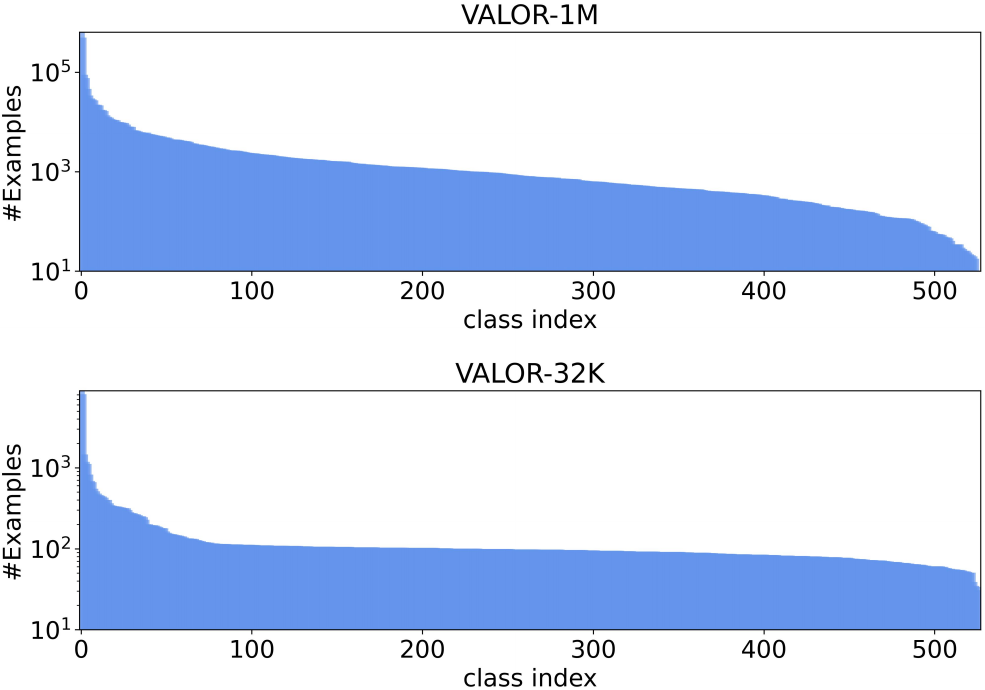}
\caption{The distributions of audio classes  in VALOR-1M and VALOR-32K.}
\label{fig:distribution}
\end{figure}

\section{VALOR DataSet for AudioVisual-Language Pretraining}

As explained in Section 2.1, video-language datasets whose captions are ASR transcriptions or alt-texts are not best choices for vision-audio-language pretraining, due to the lack of explicit correspondence between textual sentences and audio concepts. To overcome this, we propose a vision-audio-language   correlated dataset VALOR for tri-modality model pretraining and benchmarking, by annotating public audiovisual data. In the following subsections, we elaborate data collection, annotation and benchmarking process, and then analysis the characteristics of VALOR dataset.

\subsection{AudioVisual Data Collection}
Ideally, videos of  vision-audio-language dataset should contain both visual and audio tracks, with high quality and diversity. To this end, we choose videos from AudioSet\cite{gemmeke2017audio}, a large-scale dataset collected for audio event recognition. Specifically, AudioSet contains  over 2 million 10-second video clips excised from YouTube videos and each video is labeled from 527 audio classes, according to a  hierarchical ontology. It is splited into a 2M unbalanced train set, a 22k balanced train set and a 20k evaluation set.  In balanced train and evaluation set, each audio class have comparable number of videos, while the class distribution in unbalanced train set is not restricted. We downloaded videos of AudioSet whose YouTube urls are still available, filtered  low-quality broken videos, and finally achieved around 1M videos. Following \cite{gemmeke2017audio}, we split the dataset into VALOR-1M as tri-modality pretraining dataset and VALOR-32K  as audiovisual-language downstream benchmark dataset, according to audio class distributions. Specifically, Videos of VALOR-1M originate from unbalanced train set of AudioSet, and videos of VALOR-32K originate from balanced train and evaluation set of AudioSet. As Fig \ref{fig:distribution} shown,  VALOR-32K have more balanced audio class distribution compared to VALOR-1M.

\subsection{AudioVisual Caption Annotaion}

We take the paid labeling manner to  acquire audiovisual descriptions for  VALOR datasets. Considering that this annotation task is novel and more complicated than traditional video description annotation, we design a three-step interactive annotating procedure.

\textbf{Step1, annotator training.} We conduct online training for 500 annotators, emphasizing that important components like main bodies, activities, scenes, objects, and sounds should be comprehensively reflected in descriptions. Some  video-audiovisual caption pairs are provided by us to help annotators be familiar with annotation formats in advance. We also provide a dictionary that maps videoIDs to their AudioSet labels, and annotators are encouraged to  query those labels first as prior references, before  audiovisual description annotation. 

\textbf{Step2, first-stage annotation.} At this stage, we provide videos of VALOR-32K to annotators. The annotated descriptions are manually checked  by us, and we feedback  common problems and  corresponding videoIDs. Then  annotators are asked to re-annotate those unsatisfying exsamples, and build deeper understanding about annotation demands.

\textbf{Step3, second-stage annotation.} At this  stage,  annotators write audiovisual descriptions for videos of VALOR-1M. Each  description is further checked by three annotators to ensure annotation quality, and   needed to be re-annotated if more than one annotator assumed it not satisfying. The whole annotation and checking processes have taken about 2 months.

\subsection{VALOR-32K Benchmark}

Considering that current established audiovisual-language benchmarks only target at question answering (AVQA)\cite{yun2021pano,li2022learning,yang2022avqa}, we established VALOR-32K benchmark to enlarge evaluation task fields, which consists of two  tasks including audiovisual retrieval (AVR) and audiovisual captioning (AVC). As shown in Figure \ref{visual_model}, AVC demands models to generate audiovisual captions for  audiable videos and in AVR task, models are required to retrieve the most matching video candidate  according to  given audiovisual caption queries. Both AVR and AVC tasks are more challenging than existing text-to-video retrieval and video captioning tasks due to the introduction of audio modality. VALOR-32K are splited into 25K/3.5K/3.5K videos for training, validation and  testing, respectively.  The same  evaluation metrics of video retrieval and video captioning are utilized for   AVR and AVC tasks evaluation.  

\subsection{Characteristics of VALOR Dataset}

VALOR dataset is the first large-scale vision-audio-language  strong-correlated dataset, and  its biggest highlights lie in rich audio concepts and audiovisual captions.  We make quantitative and qualitative comparisons between VALOR dataset and public video-language datasets in this subsection.

\textbf{Quantitative Comparison.} To evaluate the richness of mentioned audio concepts in captions of  different datasets, we define a metric named audio concept density (ACD). We established an audio concept set according to the 632 audio classes ontology proposed by \cite{gemmeke2017audio}. Specifically, we  split one class if it contains multiple similar  concepts separated by comma, convert all words to lowercase and remove punctuations. To the end, we got 759 audio concepts. Given one caption, we preprocess it by removing punctuations and converting to lowercase, and then detect the existence of every audio concept. After iterating the whole dataset, ACD metric can be computed as follows:

\begin{equation}
  ACD =  \frac{N_{AC} }{N_{W}} 
\end{equation}
where $ N_{AC}$ equals to total number of detected audio concepts and $\ N_{W}$ is  total  number of words. 
As shown in Table \ref{data statistics}, ACD metric of VALOR dataset is much bigger than other video-language datasets. In addition, the average caption length of VALOR-1M and VALOR-32K is 16.4 and 19.8, respectively, which is much longer than other datasets like WebVid-2.5M (14.2), CC3M (10.3), thanks to  additional audio-related descriptions and high annotation quality.

\textbf{Qualitative  Comparison.} 
 We compare VALOR-1M to  ASR transcription captions based datasets like HowTo100M and  HD\_VILA\_100M,  and  alt-text captions based dataset like WebVid-2.5M. As figure \ref{fig:data representaion}  shown, captions of HowTo100M dataset are  incomplete sentences which can not  even  understood by people, let alone  vision-language correlations. Captions in HD\_VILA\_100M are more completed, but  vision-language  correlations are still weak. Specifically, the caption is transcribed from a dialog that   two people are talking about vacation recommendations, but important visual concepts like blue sky, wooden sign, and trees are not reflected in captions at all. Captions in WebVid-2.5M  has stronger visual correlation and cover more visual concepts, but they contain less audio concepts or direct descriptions about audio signal. By contrast, the annotations of VALOR focus on visual and audio clues simultaneously, reflected by the mentioned   visual concepts like black dog and sofa, and audio concepts like police alarm in the example.

\begin{figure*}[t]
\centering
\includegraphics[width=1.0\linewidth]{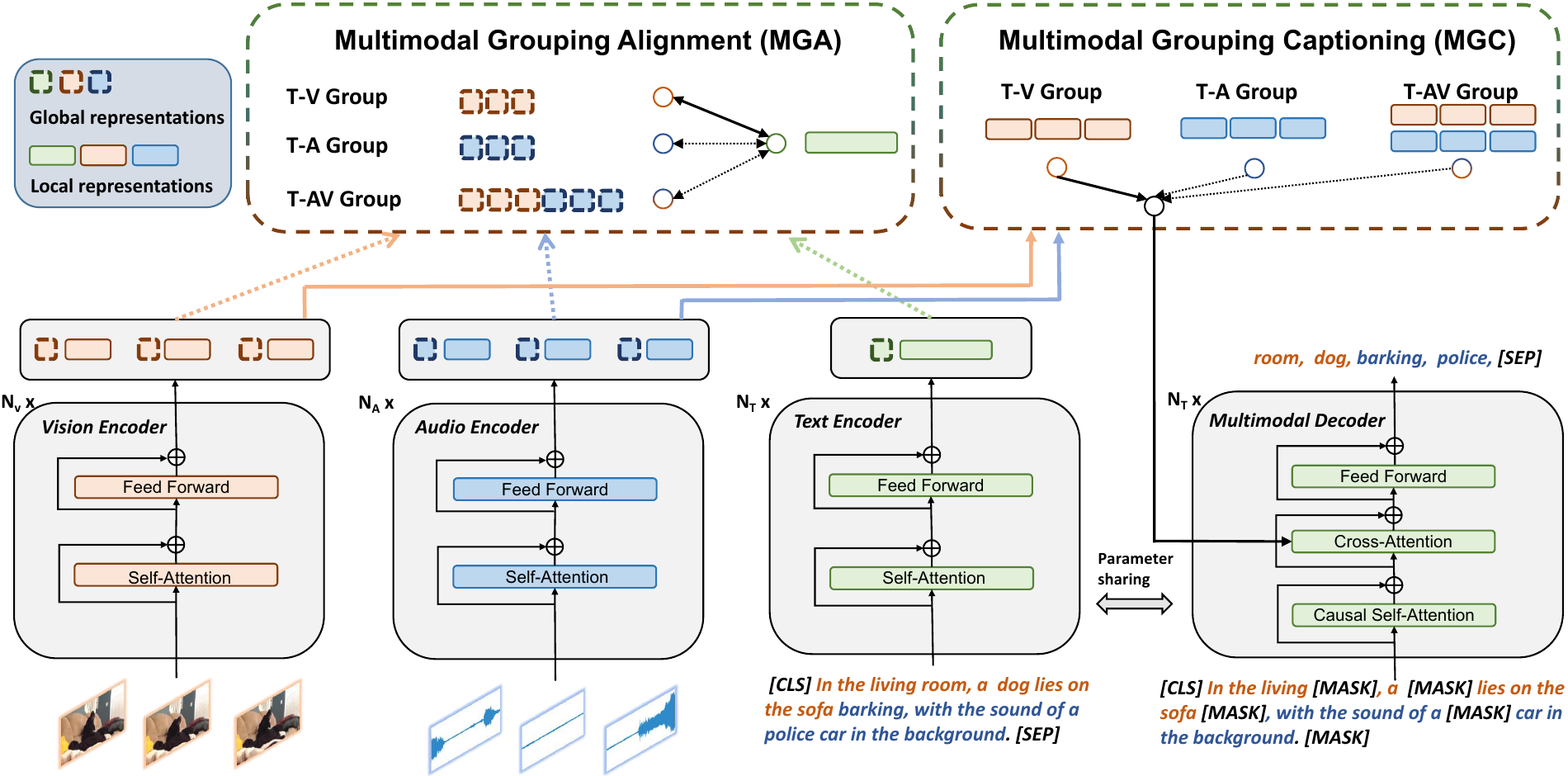}
\caption{Illustration of the overall pretraining framework of VALOR. VALOR uses three separate encoders to achieve single modality representations, and a multimodal decoder which partly shares parameters with text encoder is used for text generation. MGA and MGC tasks based on modality grouping strategy are used to improve VALOR's generalization capability to diffenent kinds of tasks and modalities.}

\label{fig:model_architecture}
\end{figure*}

\section{VALOR Model}
 We expect VALOR model to meet following demands. \uppercase\expandafter{\romannumeral1})  It can be trained fully end-to-end, avoid of pre-extracting vision or audio features, so that single modality encoders can be tuned together to learn  representations good at vision-audio-language interactions. \uppercase\expandafter{\romannumeral2}) Cross-modality alignment, discriminative and generative capabilities should be learned to improve VALOR's adaptive capability for broader cross-modality tasks.  \uppercase\expandafter{\romannumeral3}) Considering that different modalities are used in different downstream fields, VALOR should learn more generalized cross-modality  capabilities, instead of restrained into specific modality group. To this end, we made dedicate designs about model architecture and pretraining tasks, which will be  elaborated in the following subsections.

\subsection{Model Architecture}
As shown in figure \ref{fig:model_architecture}, VALOR  consists of a  text encoder, a vision encoder, an audio encoder and a multimodal decoder. This  architecture  attributes single-modality representation learning to separate encoders, whose parameters can be inherited from pretrained models to speed up convergence and improve performances.

\textbf{Text Encoder.}  BERT\cite{devlin2018bert} model is used  as text encoder. The raw sentences  are first  tokenized by BERT's tokenizer whose vocabulary size equals to  30522. The input are  summation of word embeddings and positional embeddings. The output text features are $ F_{t} \in \mathbb{R}^{N_{t}\times C_{t}}$,  where $ N_{t}$ and $ C_{t}$ are pre-defined max token length and  hidden size, respectively.

\textbf{Vision Encoder.} We have tried two vision encoders including CLIP\cite{radford2021learning}  and VideoSwin Transformer\cite{liu2022video}. Both models can take image or video singals as input. For video inputs, we sparsely sample $\rm N_{v}$ frames from a video clip, and use patch embedding layers to encode  patches.  The output feature is $ F_{v} \in \mathbb{R}^{N_{v} \times S_{v} \times C_{v}}$, where $ S_{v}$ is sequence length and $ C_{v}$ is hidden size. Frames are independently passed through CLIP encoder, while make interactions via temporal window attention in VideoSwin Transformer.  For image inputs $N_{v}$ equals to one.

\textbf{Audio Encoder.}
Audio spectrogram transformer (AST)\cite{gong2021ast,gong2022ssast} pretrained on AudioSet is used as  audio encoder. Given an audio waveform, we  split it into multiple  5 seconds long audio clips and random sample $\ N_{a}$ clips as input. Audio Clips are converted  to 64-dimensional log Mel filterbank features computed with a 25ms Hamming window every 10ms. This results in a $ 64\times 512$ spectrogram for each clip. After that the spectrograms are splited into patches, passed through patch embedding layer and fed into  audio encoder. The output feature is  $\ F_{a} \in \mathbb{R}^{N_{a} \times S_{a} \times C_{a}}$, where $ S_{a}$ is sequence length and $ C_{a}$ is hidden size.

\textbf{Multimodal Decoder.}
We use pretrained  BERT  as  multimodal decoder. A cross-attention layer is added between self-attention layer and feed-forward layer in every transformer block, whose  parameters are randomly initialized. In cross-attention layer, text feature attends to conditional features which can be the output video features, audio features or their concatenation. Except for cross-attention layers, multimodal decoder share parameters with text encoder.

\subsection{Vision-Audio-Language Cross-Modality Learning}
\label{sec4.2}
We propose Multimodal Grouping Alignment (MGA) and Multimodal 
 Grouping Captioning (MGC) tasks to conduct unified vision-audio-language learning. They are separately implemented by contrastive learning and causal masked language modeling, based on modality grouping strategy. We mainly consider three modality groups including text-vision group (T-V), text-audio group (T-A), and text-audiovisual group (T-AV), corresponding to three kinds of mainstream downstream tasks (vision-language, audio-language and audiovisual-language tasks). This strategy is necessary, imaging that only one modality group (T-AV) is learned during pretraining, the performances on vision-language and audio-language tasks will be restricted, because of the pretrain-finetune modality inconsistency.  
 
\textbf{Multimodal Grouping Alignment  (MGA).}
 We build fine-grained alignment between text and modality X via contrastive learning, and X represents  different modalities including vision (V),  audio (A) and  audiovisual (AV). Text and modality X    are considered as  positive sample if they match, and negative sample if they do not. Bi-directional contrastive loss is computed in batches to pull together positive samples and push away negative samples, which can be formulated as follows:

\begin{equation}
\begin{split}
 \ L_{MGA (T-X)}= &  -\frac{1}{2}   \sum_{i=1}^{B}{\rm log}\frac{{\rm exp} (s (T_{i}, X_{i}) /  \tau)}{\sum_{j=1}^{B}{\rm exp} (s (T_{i}, X_{j}) /  \tau)} \\
 -&  \frac{1}{2}  \sum_{i=1}^{B}{\rm log}\frac{{\rm exp} (s (T_{i}, X_{i}) /  \tau)}{\sum_{j=1}^{B}{\rm exp} (s (T_{j}, X_{i}) /  \tau)} 
\end{split}
\end{equation}
where $ s (\cdot,\cdot)$ and $\tau$ denotes similarity function and temperature, respectively.

\begin{table*}[h]
\caption{ Model configurations of VALOR$_{\rm B}$ and VALOR$_{\rm L}$. \#Example: total number of used vision-text pairs or vision-audio-text triplets. Res: resolution of images or video frames.}
\centering
\label{VALOR_model}
\scalebox{0.92}{
\begin{tabular}{lllllllll}
\toprule
Model   & Tri-modality dataset   & Dual-modality dataset & \#Example & Vision encoder & Batch size & Iteration & Params & Res  \\
\midrule
VALOR$\rm _{B}$     & VALOR-1M & WebVid-2.5M+CC3M & 6.5M                     & Video Swin$\rm _{B}$     &  512        & 200K      & 342M       & 224                   \\
VALOR$\rm _{L}$ & VALOR-1M & WebVid-2.5M+CC14M+HD\_VILA\_10M  &33.5M                  &  CLIP$\rm _{L}$      &         1024     &   500K   & 593M       & 224                   \\
\bottomrule
\end{tabular}}
\end{table*}

With regards to  similarity computation $ s (\cdot,\cdot)$, instead of directly aligning global  representations of text and modality X, we build   fine-grained  correlations between every text token and every video frame or audio clip. Specifically, we first extract global representations for each video frame and audio clip by global average pooling or using [CLS] token feature, and then tri-modality features are  projected  into the same normalized semantic space via three linear projection layers.  The normalized features are represented as   $ e_{t} \in \mathbb{R}^{N_{t} \times C}$,   $ e_{v} \in \mathbb{R}^{N_{v} \times C}$,  and  $ e_{a} \in \mathbb{R}^{N_{a} \times C}$, respectively, and C is common hidden size. The audiovisual feature $ e_{av} \in  \mathbb{R}^{ (N_{v}+N_{a}) \times C}$  is the concatenation of $ e_{v}$ and $ e_{a}$.  Then the fine-grained similarity matrix $ S_{TX} \in \mathbb{R}^{N_{t} \times N_{x}} $ is computed by dot product of $ e_{t}$ and $ e_{x}$, where $ e_{x} \in  (e_{v},e_{a},e_{av})$, and  overall  similarity is the summation of bi-directional scores, each of which is computed by maximizing $ S_{TX}$ along one matrix dimension, followed by  taking average along the other dimension.  Considering that different text tokens, visual frames or audio clips are not equally informative, we use learnable weighted average rather than equal average. The weights are achieved by feeding each modality features $ e_{t}$, $ e_{v}$ and $ e_{a}$ to independent linear layers and normalized with softmax function. The above process can be formulated as:  

\begin{equation}
\begin{split}
     s (T, X) = &  \frac{1}{2} \sum_{i=1}^{N_{t}} f_{t,\theta}^{i} (e_{t})  \max_{j=1}^{N_{x}}  ( e_{t}^{i})^{\mathrm{T}} e_{x}^{j} + \\
        &   \frac{1}{2} \sum_{j=1}^{N_{x}} f_{x,\theta}^{j} (e_{x}) \max_{i=1}^{N_{t}}  (  e_{av}^{j})^{\mathrm{T}} e_{t}^{i}
\end{split}
\label{similarity}
\end{equation}

where $ f_{\theta}$ represents the linear layers with weights  $ W \in \mathbb{R}^{C \times 1}$.   Total MGA loss is the average of three grouping alignment losses:

\begin{equation}
   L_{MGA} =   \frac{1}{3}  (L_{MGA (T-AV)} +   L_{MGA (T-V)} +   L_{MGA (T-A)})
\end{equation}

 \textbf{Multimodal Grouping Captioning  (MGC).}
Causal masked language modeling are used for this task. Specifically, input text tokens of  multimodal decoder are randomly replaced with  [MASK] tokens  with  60\% probability, and   their output features are fed into  a MLP layer  to reconstruct original tokens. In self-attention layers of multimodal decoder, causal attention mask is used to prevent information leakage and keep consistence with autoregressive inference process. Text, vision and audio features are fused through cross attention layers. Before fusion,  we first reshape $ F_{a}$ and $ F_{v}$ into two dimensions by flattening along temporal dimension, and transform them to same hidden size through linear layers, which results in $ F_{a^{'}} \in \mathbb{R}^{n_{v} \times C^{'}}$ and $ F_{v^{'}} \in \mathbb{R}^{n_{a} \times C^{'}}$, where $ n_{v}=N_{v} \times S_{v}$, $n_{a}=N_{a} \times S_{a}$ and $ C^{'}$ equals to  multimodal decoder's hidden size. The fusion audiovisual feature $ F_{av} \in \mathbb{R}^{ (n_{v}+n_{a}) \times C^{'}}$ is the concatenation of them along sequence dimension. MGC loss with modality X as condition can be formulated as:
 \begin{equation}
                    L_{MGC (T-X)} =-\mathbb{E}_{ (T,X)\in D} {\rm log} P (T_{m}|T_{< m},F_{x})
 \end{equation}
 where $\rm D$ , $ T_{m}$ and $ T_{< m}$ denote the training batch, masked token, and tokens ahead of current masked token,  respectively, and $ F_{x} \in  ( F_{v^{'}},F_{a^{'}} , F_{av})$. Total MGC loss is the average of three grouping captioning losses:

\begin{equation}
   L_{MGC} =    \frac{1}{3}  (L_{MGC (T-AV)} +   L_{MGC (T-V)} +   L_{MGC (T-A)})
\end{equation}

 In each training step, MGA and MGC is optimized simultaneously, with a tunable hypeparameter $\alpha$ to control the ratio of two tasks, so the whole training loss is formulated as :

 \begin{equation}
   L =   \alpha L_{MGA} + L_{MGC}
\end{equation}

 \subsection{Adaptation to Downstream Tasks}
 Thanks to MGA and MGC pretraining tasks introduced above, VALOR can  be easily adapted to  different types of downstream tasks and modalities. For  retrieval tasks (AVR, VR, AR), we use $ L_{MGA (T-AV)}$, $ L_{MGA (T-V)}$, $ L_{MGA (T-A)}$  as training objective, respectively, and multimodal decoder is not used. At inference time, we compute the similarity scores between each query and all candidates through Eqn. \ref{similarity}, and rank all candidates. 
 
 For captioning tasks (AVC, VC, AC), we use $ L_{MGC (T-AV)}$, $ L_{MGC (T-V)}$, $ L_{MGC (T-A)}$  as training objective, respectively. Text tokens are generated autoregressively during inference. Specifically, ``[CLS] [MASK]" is fed to  predict the first token [TK1], and ``[CLS] [TK1] [MASK]'' is fed to predict next token. The process is repeated  until [SEP] token is generated.
 
 For question answering tasks (AVQA, VQA, AQA), we formulate them  as  generative problem, so  answers can be predicted  from the whole  vocabulary instead of pre-defined top-k high frequency answer candidate sets.  During training,  MGC loss is used as training objective like captioning tasks.  Specifically, question tokens and answer tokens are concatenated to be fed into decoder, and only answer tokens are masked while question tokens are all kept  visible. The self-attention masks   in multimodal decoder are bi-directional for question tokens and causal for answer tokens. The answer inference process is also autoregressive. 
 
 \begin{table*}[htbp]
\caption{Comparison with state-of-the-art methods on VALOR-32K text-to-audiovisual retrieval benchmark and 5 text-to-video retrieval benchmarks. R@1/R@5/R@10 is reported. \#Example represents the number of used vision-text pairs or vision-audio-text triplets. Mod represents utilized modalities and V, A, S is short for vision, audio and subtitle, respectively. +DSL means that using dual softmax\cite{cheng2021improving} post processing during evaluation. Results on VALOR-32K are achieved by us using their public released codes.}

\centering
\label{sota_ret}
\scalebox{0.85}{
\begin{tabular}{lllllllll}
\toprule
Method        & \#Example & Mod  & VALOR-32K & MSRVTT         & DiDeMo         & ActivityNet    & LSMDC                    & VATEX          \\
\midrule
\multicolumn{4}{l}{\textbf{\textit{Group-A: pretrain with \textless 10M examples}}} \\

ClipBert\cite{lei2021less}      & 5.6M &V    & - & 22.0/46.8/59.9 & 20.4/48.0/60.8 &         -       &    -                          &   -             \\

Frozen\cite{bain2021frozen}        & 6.1M&V   &32.9/60.4/71.2 & 32.5/61.5/71.2 & 31.0/59.8/72.4 &    -            & 15.0/30.8/39.8  &        -        \\

BridgeFormer\cite{ge2022bridging}  & 5.5M&V   & - & 37.6/64.8/75.1 & 37.0/62.2/73.9 &          -      & 17.9/35.4/44.5 &          -       \\

MILES\cite{ge2022miles}         & 5.5M&V   &  -& 37.7/63.6/73.8 & 36.6/63.9/74.0 &          -      & 17.8/35.6/44.1 &        -         \\
OA-Trans\cite{wang2022object}      & 5.5M&V  & -  & 35.8/63.4/76.5 & 34.8/64.4/75.1 &       -         & 18.2/34.3/43.7 &    -            \\


Nagrani et al.\cite{nagrani2022learning}     & 1.03M&V+A & -& 35.8/65.1/76.9 &                &         -       &       -          &    -            \\
LF-VILA\cite{sun2022long} &8.5M         &V &-&-&35.0/64.5/75.8&35.3/65.4/-&-&- \\

\textbf{VALOR$\rm _{\textbf{B}}^{-}$(Ours)}      & 5.5M&V   & 43.3/70.3/80.0 & 36.2/64.7/75.4               &   43.2/73.9/82.4             &   37.5/67.9/80.4              &    20.0/39.1/49.0        &    59.4/90.5/95.4             \\

\textbf{VALOR$\rm _{\textbf{B}}$(Ours)}      & 6.5M&V+A   & \textbf{67.9}/\textbf{89.7}/\textbf{94.4} &  \textbf{43.0}/\textbf{72.2}/\textbf{82.1}              &   \textbf{52.2}/\textbf{80.8}/\textbf{86.8}             &   \textbf{50.5}/\textbf{79.6}/\textbf{89.1}             &  \textbf{25.1}/\textbf{45.8}/\textbf{55.2}           &    \textbf{67.5}/\textbf{94.1}/\textbf{97.4}              \\
\midrule
\multicolumn{4}{l}{\textbf{\textit{Group-B: pretrain with \textgreater 10M examples or inherit CLIP model weights}}} \\
SINGULARITY\cite{lei2022revealing}   & 17M&V    & - & 41.5/68.7/77.0 & 53.9/79.4/86.9 & 47.1/75.5/85.5 &      -          &         -            \\
LAVENDER\cite{li2022lavender}      & 30M&V    & - & 40.7/66.9/77.6 & 53.4/78.6/85.3 &          -      & 26.1/46.4/57.3 &           -    \\

MV-GPT\cite{seo2022end}        & 53M&V+S  & - & 37.3/65.5/75.1 &         -       &           -     &            -    &        -            \\
TACo\cite{yang2021taco}          & 136M &V   & - & 28.4/57.8/71.2 &       -         & 30.4/61.2/-    &      -          &            -          \\
Support-set\cite{patrick2020support}   & 136M & V & -   & 30.1/58.5/69.3 &         -       & 29.2/61.6/-    &        -        &  44.9/82.1/89.7 \\
MMT\cite{gabeur2020multi}           & 136M & V+A  & -& 26.6/57.1/69.6 &           -     & 28.7/61.4/-    & 12.9/29.9/40.1 &        -          \\
AVLNet\cite{rouditchenko2020avlnet}        & 136M & V+A  & 21.6/47.2/59.8& 22.5/50.5/64.1 &          -      &   -             & 11.4/26.0/34.6 &          -             \\
Gabeur et al.\cite{gabeur2022masking}      & 136M&V+A+S & -& 28.7/59.5/70.3 &       -         & 29.0/61.7/-    &          -            &   -             \\
All-in-one\cite{wang2022all}   & 138M&V    & -& 37.9/68.1/77.1 & 32.7/61.4/73.5 & 22.4/53.7/67.7 &         -          &        -        \\
VIOLET\cite{fu2021violet}        & 186M&V   &  -& 34.5/63.0/73.4 & 32.6/62.8/74.7 &          -      &  16.1/36.6/41.2      &    -            \\
CLIP4Clip\cite{luo2022clip4clip}     & -&V   & 43.4/69.9/79.7 & 44.5/71.4/81.6 & 43.4/70.2/80.6 & 40.5/72.4/-    & 22.6/41.0/49.1 &  55.9/89.2/95.0 \\

TS2-Net\cite{liu2022ts2}       & -  & V  &-& 49.4/75.6/85.3 & 41.8/71.6/82.0 & 41.0/73.6/84.5 & 23.4/42.3/50.9       & 59.1/90.0/95.2 \\

X-CLIP\cite{ma2022x}  & - &V &- &49.3/75.8/84.8&47.8/79.3/-&46.2/75.5/-&26.1/48.4/46.7&- \\

ECLIPSE\cite{lin2022eclipse} &- &V+A             & -&-&44.2/-/-&45.3/75.7/86.2&-&- \\
DCR\cite{wang2022disentangled}          & -&V    & -& 50.2/76.5/84.7 & 49.0/76.5/84.5 & 46.2/77.3/88.2 & 26.5/47.6/56.8  & 65.7/92.6/96.7 \\
HunYuan\_tvr+DSL\cite{min2022hunyuan_tvr} & -&V  &  - & 55.0/80.4/86.8 & 52.1/78.2/85.7&  57.3/84.8/93.1 &  29.7/46.4/55.4  &       -         \\

CLIP-VIP+DSL \cite{xue2022clip} & 100M&V    & -& 57.7/80.5/88.2 & 55.3/82.0/89.3&  61.4/85.7/92.6 &  30.7/51.4/60.6 &        -         \\

InternVideo+DSL\cite{wang2022internvideo}  & 147.6M &V &- &55.2/-/-&57.9/-/-&62.2/-/-&34.0/-/-&71.1/-/- \\

\textbf{VALOR$\rm _{\textbf{L}}$(Ours)}     & 33.5M&V+A & 73.2/91.6/95.4 & 54.4/79.8/87.6             &    57.6/83.3/88.8           &    63.4/87.8/94.1          &    31.8/52.8/62.4                       &      76.9/96.7/98.6           \\

\textbf{VALOR$\rm _{\textbf{L}}$+DSL(Ours)}     & 33.5M&V+A &\textbf{80.9}/\textbf{93.9}/\textbf{97.1} & \textbf{59.9}/\textbf{83.5}/\textbf{89.6}             &    \textbf{61.5}/\textbf{85.3}/\textbf{90.4}            &    \textbf{70.1}/\textbf{90.8}/\textbf{95.3}           &    \textbf{34.2}/\textbf{56.0}/\textbf{64.1}                     &      \textbf{78.5}/\textbf{97.1}/\textbf{98.7}           \\
\bottomrule
\end{tabular}}
\end{table*}

\section{Experiments}
In this section, we first introduce basic experiment settings including  pretraining datasets, downstream benchmarks and implementation details. After that we compare VALOR to state-of-the-art methods on various of benchmarks. Finally, we present detailed ablation studies to demonstrate  effectiveness of proposed method and visualize VALOR's prediction results.

\subsection{Experiment Settings}
\subsubsection{Pretraining Datasets}
The following 4 datasets are used for VALOR's pretraining.
\textbf{VALOR-1M} is the proposed  tri-modality dataset, which contains one  million open-domain audiable videos with manually annotated audiovisual captions. 

\textbf{WebVid-2.5M}\cite{bain2021frozen} is a web-crawled dataset which contains about 2.5M videos paired with alt-texts. Recently its larger version,  WebVid-10M is also released, but is not utilized in this work. 

\textbf{CC14M} is a combination of  series of image-language datasets including MSCOCO\cite{lin2014microsoft}, Visual Genome\cite{krishna2017visual}, SBU\cite{ordonez2011im2text}, CC3M\cite{sharma2018conceptual} and CC12M\cite{changpinyo2021conceptual}, leading to total 14M images or 20M image-text pairs. We exclude SBU dataset due to that   too much images are invalid when downloading. 

\textbf{HD\_VILA\_100M}\cite{xue2022advancing} is a high resolution open-domain video-text datasets. It consists of 100M videos with ASR transcriptions. Due to storage limitation,  we only use a randomly sampled 10M videos subset (HD\_VILA\_10M).


\subsubsection{Downstream Tasks}

For retrieval  tasks, we evaluate VALOR on  9 public datasets including  VR (MSRVTT\cite{xu2016msr}, DiDeMo\cite{anne2017localizing}, LSMDC\cite{rohrbach2017movie}, ActivityNet\cite{krishna2017dense}, VATEX\cite{wang2019vatex} and MSCOCO\cite{lin2014microsoft}), AR (ClothoV1\cite{drossos2020clotho}  and AudioCaps\cite{kim2019audiocaps}) and AVR (proposed VALOR-32K). For DiDeMo and ActivityNet datasets, we follow other works to concatenate multiple short  temporal descriptions into  long sentences, and evaluate paragragh-to-video retrieval. Recall at rank K  (R@K, K=1,5,10) are used as metrics.

For captioning tasks, we evaluate VALOR on 7  public datasets including VC (MSVD\cite{chen2011collecting}, MSRVTT, VATEX and MSCOCO), AC (ClothoV1 and AudioCaps) and AVC (proposed VALOR-32K). BLEU4 (B4)\cite{papineni2002bleu}, METEOR (M)\cite{banerjee2005meteor}, ROUGE-L (R)\cite{banerjee2005automatic}, CIDEr (C)\cite{vedantam2015cider} and SPICE (S)\cite{anderson2016spice} are used as metrics. During inference, beam search is used and beam size is 3.

For open-ended question answering tasks, we evaluate on 6 public datasets including VQA (MSVD-QA\cite{xu2017video}, MSRVTT-QA\cite{xu2017video}, ActivityNet-QA\cite{yu2019activitynet}, TGIF-Frame QA\cite{jang2017tgif}, VQAv2\cite{goyal2017making}) and AVQA (MUSIC-AVQA\cite{li2022learning}). Accuracy is used as metric. During inference, we use greedy search to generate answers from whole vocabulary with no restrictions.

\begin{table*}[t]
\centering
\caption{Comparison with state-of-the-art methods on VALOR-32K audiovisual captioning benchmark and 3 video  captioning benchmarks. Given that most methods use reinforcement learning method\cite{rennie2017self} to improve model's performance on VATEX dataset, we also follow them for fair comparison, and corresponding results are marked with *. Results on VALOR-32K are achieved by us using their public released codes.}
\label{sota_cap}
\scalebox{0.85}{
\begin{tabular}{lllllllllllllllllll}
\toprule
 \multirow{2}{*}{Method}               & \multirow{2}{*}{\#Example} &   \multirow{2}{*}{Mod}     & \multicolumn{4}{c}{VALOR-32K}    & \multicolumn{4}{c}{MSVD}      & \multicolumn{4}{c}{MSRVTT}                                       & \multicolumn{4}{c}{VATEX}                                        \\
\cmidrule (lr){4-7}  \cmidrule (lr){8-11} \cmidrule (lr){12-15} \cmidrule (lr){16-19}
 & &&  B@4   & M     & R     & C &B@4   & M     & R     & C     & B@4           & M              & R              & C              & B@4           & M              & R              & C              \\
\midrule
\multicolumn{4}{l}{\textbf{\textit{Group-A: pretrain with \textless 10M examples}}} \\
ORG-TRL\cite{zhang2020object}         & - &V    &-&-&-&-   & 54.3  & 36.4  & 73.9  & 95.2  & 43.6          & 28.8           & 62.1           & 50.9           & 32.1          & 22.2           & 48.9           & 49.7           \\
OpenBook\cite{zhang2021open}\        & - &V    &-&-&-&-    &   -    &    -   &    -   &    -   & 42.8          & 29.3           & 61.7           & 52.9           & 33.9          & 23.7           & 50.2           & 57.5           \\
SwinBERT\cite{lin2022swinbert}        & - &V    &5.4&10.7&27.2&27.3    & 58.2  & 41.3  & 77.5  & 120.6 & 41.9          & 29.9           & 62.1           & 53.8           & 38.7          & 26.2           & 53.2           & 73.0           \\
SMPFF\cite{chen2021mm21}          & - &V+A    &7.5&12.6&28.6&37.1  &  -     &   -    &     -  &     -  & 48.4         & 30.6          & 64.9          & 58.5          & 39.7         & 26.0          & 53.6          & 70.5          \\

VIOLETv2\cite{fu2022empirical}          & 5.5M &V    &-&-&-&-  & -     & -     & -     & 139.2 & -             & -              & -              & 58.0           &  -             & -               &       -         &  -              \\

\textbf{VALOR$\rm _{\textbf{B}}^{-}$(Ours)}          & 5.5M &V    &8.0&13.5&29.4&44.3  &     74.3 & 47.1     & 83.8     & 156.1 & 48.1             & 30.4              & 64.3              & 61.5           &  40.7            & 26.1               &       53.8         &  71.6              \\

\textbf{VALOR$\rm _{\textbf{B}}$(Ours)}        & 6.5M&V+A   &\textbf{8.9}&\textbf{14.8}&\textbf{30.8}&\textbf{55.7}   & \textbf{76.1}      &   \textbf{48.0}    &  \textbf{85.2}     &\textbf{162.1}       &   \textbf{53.8}            &     \textbf{32.3}           &      \textbf{67.0}          &    \textbf{66.6}            &  \textbf{41.9}            & \textbf{26.6}               &  \textbf{54.6}              &  \textbf{73.9}              \\
\midrule
\multicolumn{4}{l}{\textbf{ \textit{Group-B: pretrain with \textgreater 10M examples}}} \\
LAVENDER\cite{li2022lavender}        & 30M &V     &-&-&-&-  & -     & -     & -     & 150.7 & -             & -              & -              & 60.1           &    -           &   -             &          -      &       -         \\

Support-set\cite{patrick2020support}     & 136M &V   &-&-&-&-   &   -    &     -  &    -   &   -    & 38.9          & 28.2           & 59.8           & 48.6           & 32.8          & 24.4           & 49.1           & 51.2           \\
VALUE\cite{li2021value}           & 136M &V+S   &-&-&-&- &   -    &   -    &  -     &   -    &          -     &    -            &     -           &    -            & 32.9          & 24.0           & 50.0           & 58.1           \\
MV-GPT\cite{seo2022end} & 136M &V+S   &-&-&-&- &    -   &  -     &    -   &     -  & 48.9          & 38.7           & 64.0           & 60.0           &   -            &   -             &         -       &     -           \\
GIT$\rm _{L}$\cite{wang2022git}       & 20M&V    &-&-&-&-   & 75.8  & 48.7  & 85.5  & 162.9 & 48.7          & 30.9           & 64.9           & 64.1           & 41.6*         & 26.2*          & 54.3*          & 72.5*          \\
GIT\cite{wang2022git}        & 800M&V   &-&-&-&-   & 79.5  & \textbf{51.1}  & 87.3  & \textbf{180.2} & 53.8          & 32.9           & 67.7           & 73.9           & 41.6*         & 28.1*          & 55.4*          & 91.5*          \\
{GIT2 (5.1B)}\cite{wang2022git} & 12.9B&V  &-&-&-&- & 82.2  & 52.3  & 88.7 & 185.4 & 54.8          & 33.1           & 68.2          & 75.9          & 42.7*        & 28.8*          & 56.5*         & 94.5*          \\

\textbf{VALOR$\rm _{\textbf{L}}$(Ours)}       & 33.5M&V+A   &\textbf{9.6}&\textbf{15.4}&\textbf{31.8}&\textbf{61.5} & 80.7  & 51.0 & 87.9  & 178.5 & 54.4 &      32.9     &   68.0         &    74.0       & \textbf{45.6*} &   \textbf{29.4*}         &         \textbf{57.4*}  &      \textbf{95.8*}      \\

\bottomrule
\end{tabular}}
\end{table*}

\begin{table*}[]
\caption{Comparison with state-of-the-art methods on 5 open-ended video QA and audioviusal QA benchmarks.}
\label{sota_qa}
\centering
\scalebox{0.85}{
\begin{tabular}{lllccccc}
\toprule
Method        & \#Example & Mod & MSRVTT-QA        & MSVD-QA          & TGIF-FrameQA    & ActivityNet-QA &MUSIC-AVQA \\
\midrule
\multicolumn{4}{l}{\textbf{\textit{Group-A: pretrain with \textless 10M examples}}} \\
QueST\cite{jiang2020divide}         & - &V       & 34.6          & 34.6          & 59.7          &    -   &    -       \\
MUSIC-AVQA\cite{li2022learning} & - &V+A &-&-&-&-&71.5 \\
ClipBERT\cite{lei2021less}      & 5.6M&V     & 37.4          & -             & 60.3          &     -     &    -    \\

VIOLET\cite{fu2022empirical}        & 5.5M&V     & 44.5          & 54.7          & 72.8          &     -   &    -      \\

Clover\cite{huang2022clover}        & 5.5M&V     & 43.9          & 51.9          & 71.4          &    -   &    -       \\
\textbf{VALOR$\rm _{\textbf{B}}^{-}$(Ours)}      & 5.5M&V     &     44.5         &54.9               &      73.0       &  43.7     &  74.8        \\
\textbf{VALOR$\rm _{\textbf{B}}$(Ours)}      & 6.5M&V+A     &  \textbf{46.7}             &   \textbf{56.4}            & \textbf{74.5}              &   \textbf{44.8}    &    \textbf{76.6}      \\
\midrule
\multicolumn{4}{l}{\textbf{\textit{Group-B: pretrain with \textgreater 10M examples}}} \\
SINGULARITY\cite{lei2022revealing}   & 17M&V      & 43.5          &          -     &     -          & 43.1   &    -      \\
LAVENDER\cite{li2022lavender}      & 30M&V      & 45.0          & 56.6          & 73.5          &  -    &    - \\
JustAsk\cite{yang2021just}       & 69M&V      & 41.5          & 46.3          & -             & 38.9     &    -    \\
MV-GPT\cite{seo2022end}        & 53M&V+S    & 41.7          & -             & -             & 39.1       &    -  \\
MERLOT\cite{zellers2021merlot}        & 180M&V     & 43.1          & -             & 69.5          & 41.4     &    -    \\
All-in-one\cite{wang2022all}    & 228.5M&V   & 46.8          & 48.3          & 66.3          &       -    &    -   \\

Flamingo (80B)\cite{alayrac2022flamingo} & 2.3B&V     & 47.4          &          -     &        -       &   -   &    -        \\
FrozenBiLM\cite{yang2022zero}    & 10M&V     & 47.0          & 54.8          & 68.6          & 43.2     &    -    \\

InternVideo\cite{wang2022internvideo} &147.6M &V & 47.1        &55.5 &72.2&-&- \\

VideoCoCa (2.1B)\cite{yan2022video} &4.8B &V & 46.0        &56.9 &-&-&- \\
GIT$\rm _{L}$\cite{wang2022git}     & 20M&V     & 42.7          & 55.1          & 71.9          &     -     &    -    \\
GIT\cite{wang2022git}      & 800M&V     & 43.2          & 56.8          & 72.8          &       -    &    -   \\
GIT2 (5.1B)\cite{wang2022git}         & 12.9B &V   & 45.6        & 58.2         & 74.9          &    -      &    -    \\

\textbf{VALOR$\rm _{\textbf{L}}$(Ours)}     & 33.5M&V+A & \textbf{49.2} & \textbf{60.0} & \textbf{78.7} & \textbf{48.6}  &    \textbf{78.9}  \\
\bottomrule
\end{tabular}}
\end{table*}

\subsubsection{Implementation Details}
All models  are trained based on PyTorch framework and   8 Tesla A100 cards. The pretraining learning rate is 1e-4. Warm up and linear learning rate decay scheduler is used. For ablation studies, unless specially explained, we use Video Swin Transformer-small pretrained on Kinetics-400 as vision encoder. We pretrain on VALOR-1M for 4 epoch with  512  batch size. 

For state-of-the-arts comparison, we train two models with different scales, namely VALOR$\rm _{B}$ and VALOR$\rm _{L}$ whose  specific configurations are presented in Table \ref{VALOR_model}.  Compared to VALOR$\rm _{B}$, VALOR$\rm _{L}$ is trained with more training data, larger batch size,  more iterations, and  use  more powerful vision encoder. Except for different vision encoders, both model use the same text/multimodal encoder (BERT$\rm _{B}$) and  audio encoder (AST). At each iteration, we sample a dataset according to pre-defined weights, and if a dual-modality dataset is sampled, no audio is used. For each video, we sample  1 video frame and 1 audio clip during pretraining. During finetuning, we use task-specfic learning rate and sample numbers.

\subsection{Comparison to State-of-the-arts}

\subsubsection{Video-Language Benchmarks}

\textbf{Text-to-Video Retrieval.} As shown in Table \ref{sota_ret}, VALOR$\rm _{B}$ outperforms all models in Group-A with evident gaps on VALOR-32K, MSRVTT, DiDeMo  and LSMDC datasets. On ActivityNet and VATEX datasets,  VALOR$\rm _{B}$  even surpasses all models in Group-B, with only 6.5M pretraining data, which demonstrates high effectiveness and efficiency of VALOR. We also train a base-level model using only WebVid-2.5M and CC3M and without involving audio in both pretraining and finetuning, which is denoted as VALOR$\rm _{B}^{-}$. From the comparison between  VALOR$\rm _{B}$ and  VALOR$\rm _{B}^{-}$ we can find VALOR-1M dataset and audio modality vitals for VALOR's high performance. In addition, compared with models in Group-B,  VALOR$\rm _{L}$  achieves new SOTA results on MSRVTT, DiDeMo, ActivityNet, LSMDC, VATEX datasets, and outperforms previous SOTA performances (R@1) by 3.8\%, 6.2\%, 12.7\%, 0.6\%,  10.4\%, respectively. We attribute VALOR's huge improvements to \romannumeral1)  vision-audio-language alignment construction instead of dual modality alignment. \romannumeral2)  fine-grained alignment construction between text and audiovisual signals instead of coarse-grained alignment.

\begin{table}[t]
\centering
\caption{Comparison with state-of-the-art methods on 4 audio-language benchmarks.}
\label{sota_audio}
\scalebox{0.9}{
\begin{tabular}{c}

 (a) Text-to-Audio Retrieval \\
\begin{tabular}{lllllll}
\toprule
\multirow{2}{*}{Method} & \multicolumn{3}{c}{ClothoV1} & \multicolumn{3}{c}{AudioCaps} \\
\cmidrule (lr){2-4} \cmidrule (lr){5-7}
                        & R@1     & R@5     & R@10   & R@1      & R@5      & R@10    \\
\midrule
Oncescu et al.\cite{oncescu2021audio}          & 9.6     & -       & 40.1   & 25.1     & -        & 73.2    \\
Nagrani et al.\cite{nagrani2022learning}           & 12.6    & -       & 45.4   & 35.5     & -        & \textbf{84.5}    \\

\textbf{VALOR$\rm _{\textbf{B}}$}                    & \textbf{17.5}    & \textbf{42.7}    & \textbf{55.3}   & \textbf{40.1}     & \textbf{73.9}     & 83.1    \\

\bottomrule
\end{tabular} \\
\\

 (b) Audio Captioning \\
\scalebox{0.9}{
\begin{tabular}{lllllllll}
\toprule
\multirow{2}{*}{Method} & \multicolumn{4}{c}{ClothoV1} & \multicolumn{4}{c}{AudioCaps}                                  \\
\cmidrule (lr){2-5} \cmidrule (lr){6-9}
                        & B@4   & M    & R    & C    & B@4  & M    & R      & C                                       \\
\midrule
Kim et al.\cite{kim2019audiocaps}              &    -   &   -   &    -  &   -   & 21.9 & 20.3 & 45.0 & 59.3                                  \\

Xu et al.\cite{xu2020crnn}               & 15.6  & 16.2 & 36.8 & 33.8 &    -  &  -    &-        &  -                                       \\
Chen et al.\cite{chen2020audio}             & 15.1  & 16.0 & 35.6 & 34.6 &   -   &  -    & -       &   -                                      \\
Xu et al.\cite{xu2021investigating}               & 15.9  & 16.9 & 36.8 & 37.7 & 23.1 & 22.9 & 46.7 & 66.0 \\
Koh et al.\cite{koh2022automated}              & 16.8  & 16.5 & 37.3 & 38.0 &   -   &   -   & -       & -                                        \\
ACT\cite{mei2021audio}                     &    -   &   -   &   -   &   -   & 25.2 & 22.2 & 46.8   & 67.9                                    \\
Liu et al.\cite{liu2022leveraging}              &    -   &  -    &  -    &  -    & 25.1 & 23.2 & 48.0   & 66.7                                    \\

\textbf{VALOR$\rm _{\textbf{B}}$}                  & \textbf{16.2}  & \textbf{17.4} & \textbf{38.2} & \textbf{42.3} & \textbf{27.0} & \textbf{23.1} & \textbf{49.4}   & \textbf{74.1}                                    \\

\bottomrule
\end{tabular}} \\
\\

\end{tabular}}
\end{table}

It is noted that VALOR also outperforms methods which additionally utilize audio, subtitle or both modalities\cite{nagrani2022learning,rouditchenko2020avlnet,gabeur2020multi,lin2022eclipse,seo2022end}, demonstrating the effectiveness of fine-grained tri-modality alignment modeling in VALOR, and also the importance of utilizing strong-correlated tri-modality pretraining data. Different from short-form video retrieval datasets (\textless30s) like MSRVTT and LSMDC, long-form datasets (\textgreater1min) including DiDeMo and ActivityNet are more challenging due to more complicated temporal relationship between long videos and paragraghs.   VALOR significantly outperforms methods that specializes in long-form video retrieval\cite{sun2022long,lin2022eclipse}  without bells and whistles, which has shown VALOR's powerful generalization capabilities given that VALOR only  saw  short videos (around 10s) during pretraining.
In addition, compared with  methods\cite{li2022lavender,lei2022revealing} who train models with video-text matching (VTM) loss, VALOR possesses higher inference efficiency and performance at the same time.

 \textbf{Video Captioning.} As presented in Table \ref{sota_cap},  VALOR$\rm _{B}$ outperforms  all models in Group-A on  4 benchmarks. In Group-B, we mainly compare VALOR to GIT model\cite{wang2022git}, a recently proposed  large-scale generative pretraining model which has achieved SOTA results on many vision captioning benchmarks. Specifically, GIT  has four scales, named $\rm GIT_{B}$, GIT$\rm _{L}$, GIT and GIT2 according to different parameter and data size. It is noted that   GIT$\rm _{L}$  uses comparable amount of training data and the same vision encoder as VALOR$\rm _{L}$  (i.e., CLIP$\rm _{L}$), while  GIT uses a bigger vision encoder  (CoSwin model pretrained by Florence\cite{yuan2021florence} and larger data size. GIT2 even uses a 4.8B DaViT\cite{ding2022davit} as vision encoder and  12.9B vision-text pairs as training data. From comparison results we can find that  VALOR$\rm _{L}$ outperforms GIT$\rm _{L}$ and GIT on most metrics of all three benchmarks with huge margins. In addition, VALOR$\rm _{L}$ even outperforms GIT2 on VATEX benchmarks, with much smaller parameters (11.6\%), data size (0.26\%) and image resolution (224 vs 384). These results  demonstrate that learning audiovisual conditioned text generation  (scaling up pretraining model from modality dimension) is  more efficient and effective  compared with scaling up from model parameter and data size dimensions.

\textbf{Open-Ended Video QA.}  As tabel \ref{sota_qa} shows, VALOR$\rm _{B}$ outperforms  all models in Group-A on  five benchmarks.  In  Group-B, FrozenBiLM uses the same vision encoder as VALOR$\rm _{L}$ (i.e., CLIP$\rm _{L}$), and a more powerful decoder  (a 890M DeBERTa-V2-XLarge model\cite{he2020deberta}). Flamingo has 135$\times$ parameters and 68.7$\times$ training data than VALOR$\rm _{L}$. VideoCoCa inherited weights form CoCa which has 3.5$\times$ parameters and 143.3$\times$ training data than VALOR$\rm _{L}$. GIT2 has 8.6$\times$ parameters and 382.1$\times$ training data than VALOR$\rm _{L}$. Even with much smaller parameters and training data,  VALOR$\rm _{L}$ achieves new SOTA performances on MSRVTT-QA, MSVD-QA, TGIF-FrameQA, ActivityNet-QA benchmarks, and surpasses  previous SOTA methods   by 3.8\%, 3.4\%, 5.1\%, 12.5\%, respectively.  On audiovisual question answering benchmark MUSIC-AVQA, VALOR$\rm _{B}$ and VALOR$\rm _{L}$ improves the baseline by 7.1\% and 10.3\%, respectively.

\subsubsection{Audio-Language Benchmarks}
  As shown in Table \ref{sota_audio}, with regards to text-to-audio retrieval task, VALOR achieves new sota performances on ClothoV1, AudioCaps benchmarks, and outperforms previous SOTA methods  (R@1) by 38.9\%, 13.0\%, respectively.   Nagarani's method pretrains tri-modality model on their proposed VideoCC3M dataset\cite{nagrani2022learning}, which is  achieved by collecting videos from InterNet which have high similarities to CC3M's images, and directly take CC3M's captions as video captions. By contrast, VALOR achieves evidently better performances on two text-to-audio benchmarks, thanks to  that  audio-language correlations in VALOR-1M are much more explicit and stronger than those in VideoCC3M. In addition,  their method needs to separately pretrain two models for video and  audio retrieval, while we can finetune a single pretaining model on video, audio and audiovisual retrieval tasks, thanks to proposed modality grouping strategy. With regards to audio captioning benchmarks, compared with models directly training on target datasets, VALOR shows better results  and achieves new SOTA results on two benchmarks.

\begin{table}[]
\caption{Comparison with state-of-the-art methods on 3 image-language benchmarks. Results marked with * indicate that they  are achieved by using reinforcement learning. VALOR takes 392 as image resolution on all three benchmarks.}
\label{sota_image}
\scalebox{0.73}{
\begin{tabular}{lllllllll}
\toprule
\multirow{2}{*}{Method}          & \multirow{2}{*}{\#Example} & \multicolumn{3}{c}{COCO-Retrieval} & \multicolumn{2}{c}{COCO-Caption} & \multicolumn{2}{c}{VQA v2} \\
\cmidrule (lr){3-5} \cmidrule (lr){6-7} \cmidrule (lr){8-9}

                &                  & R@1               & R@5     & R@10    & C               & S              & dev         & std         \\
\midrule
UNITER\cite{chen2020uniter}          & 10M              & 52.9             & 79.9   & 88.0   &   -             &         -       & 73.82       & 74.02       \\
Oscar\cite{li2020oscar}           & 10M              & 57.5             & 82.8   & 89.8   & 140.9*          & 25.2*          & 73.61       & 73.82       \\
UFO\cite{wang2021ufo}             & 10M              & 59.2             & 83.6   & 90.5   & 131.2           & 23.3           & 76.64       & 76.76       \\
VinVL\cite{zhang2021vinvl}           & 10M              & 58.8             & 83.5   & 90.3   & 140.0*          & 24.5*          & 76.52       & 76.60       \\
ALBEF\cite{li2021align}           & 20M              & 60.7             & 84.3   & 90.5   &         -        &     -           & 75.84       & 76.04       \\
METER\cite{dou2022empirical}           & 10M              & 57.9             & 82.7   & 90.1   &            -     &     -           & 77.68       & 77.64       \\
ALIGN\cite{jia2021scaling}           & 1.8B             & 59.9             & 83.3   & 89.8   &            -     &       -         &  -           &           -  \\
FILIP\cite{yao2021filip}           & 340M             & 61.2             & 84.3   & 90.6   &       -          &      -          & -            &        -     \\
Florence\cite{yuan2021florence}        & 900M             & 63.2             & 85.7   & -      &    -             &       -         & 80.16       & 80.36       \\
BLIP\cite{li2022blip}            & 135M             & 65.1             & 86.3   & 91.8   & 136.7           &  -              & 78.25       & 78.32       \\

Flamingo (80B)\cite{alayrac2022flamingo}   & 2.3B             &       -           &  -      &   -     & 138.1           &     -           & 82.0        & 82.1        \\
LEMON\cite{hu2022scaling}           & 200M             &         -         &    -    &    -    & 145.5*          & 25.5*          &  -           &       -      \\
SimVLM\cite{wang2021simvlm}          & 1.8B             &             -     &     -   &    -    & 143.3           & 25.4           & 80.03       & 80.34       \\
CoCa\cite{yu2022coca}            & 4.8B             &        -          &   -     &       - & 143.6           & 24.7           &   82.3          & 82.3            \\
GIT$\rm _{L}$\cite{wang2022git}       & 20M             &     -             &  -      & -       & 144.6*          & 25.4*          & 75.5        & -           \\
GIT\cite{wang2022git}        & 800M             &       -           &  -      &     -   & 151.1*          & 26.3*          & 78.6        & 78.8        \\
GIT2 (5.1B)\cite{wang2022git} & 12.9B            &           -       &    -    &   -     &  152.7*         & 26.4*          & 81.7        & 81.9        \\
PALI (16.9B)\cite{wang2022git} & 1.6B            &           -       &    -    &   -     &   149.1        & -          & 84.3        & 84.3       \\
\textbf{VALOR$\rm _{\textbf{L}}$}        & 33.5M           & 61.4   &   84.4      &  90.9      & 152.5*          & 25.7*          &    78.46         & 78.62            \\
\bottomrule
\end{tabular}}
\end{table}

\begin{figure*}[ht]
\centering
\includegraphics[width=1.0\linewidth]{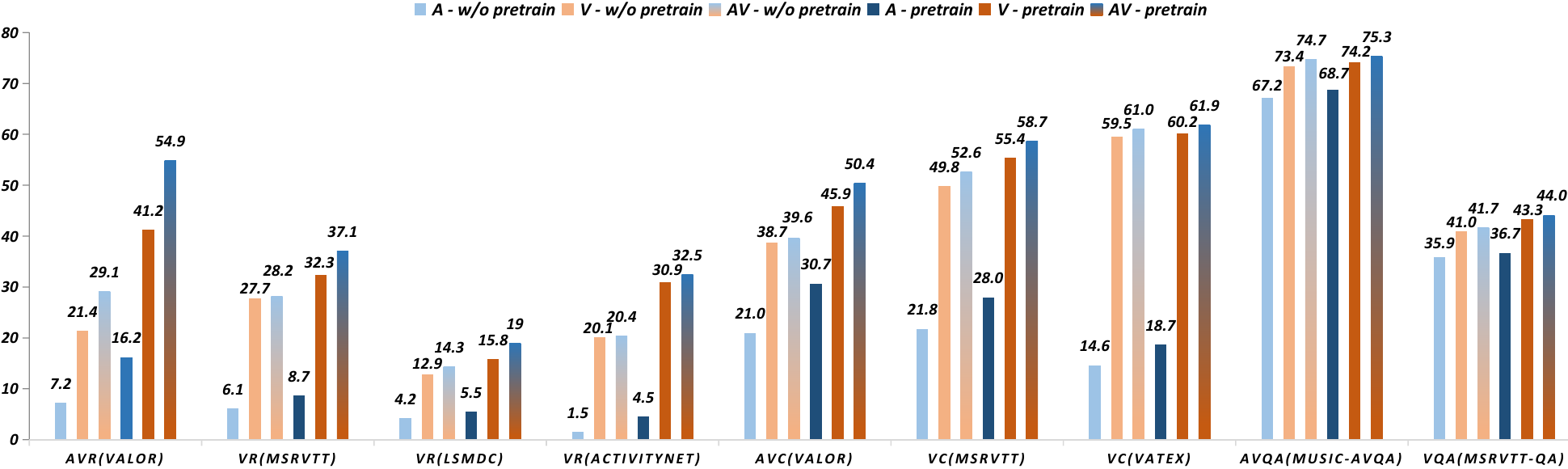}
\caption{Experiment results of using audio (A), vision (V) or both modalities(AV)  for 9 benchmarks of 6 tasks. W or w/o pretraining denotes whether the model is pretrained on VALOR-1M  or not.  R@1, CIDEr and  Acc metrics are reported for retrieval, captioning and QA tasks, respectively.}
\label{fig:audio_eff}
\end{figure*}

\begin{table*}[h]
\centering
\caption{Downstream  performances of models pretrained on VALOR-1M with  different modality groups in MGA task. Only $ L_{MGA}$ is used.  Zero-shot R@1 / finetune R@1 are reported.}
\label{MGAA}
\scalebox{0.88}{
\begin{tabular}{ccccccccccccccc}
\toprule
\multirow{2}{*}{Name}& \multicolumn{6}{c} {Pretraining Modality Groups} & \multicolumn{2}{c}{T-V} & \multicolumn{2}{c}{T-A} & T-AV      & V-A       & A-TV      & V-TA      \\

\cmidrule (lr){2-7}\cmidrule (lr){8-9} \cmidrule (lr){10-11} \cmidrule (lr){12-15}

&T-V&T-A&T-AV&V-A&A-TV&V-TA        & VALOR-32K     & MSVD       & VALOR-32K      & ClothoV1    & \multicolumn{4}{c}{VALOR-32K}    \\
\midrule
M1&\Checkmark&&&&&                                & \textbf{38.3}/\textbf{42.7}  & \textbf{30.7}/36.0  & 0.0/7.2     & 0.2/9.0   & 38.3/46.9 &       -    &    -       &    -       \\
M2&&\Checkmark&&&&                                & 0.0/18.1   & 0.2/14.7   & 17.1/16.7   & 8.4/16.5  & 14.5/30.8 &   -        &        -   &  -         \\
M3&&&\Checkmark&&&                                & 34.9/40.1  & 28.7/34.8  & 10.2/12.3   & 8.1/14.7  & 50.5/55.4 &     -      &       -    &  -         \\
M4&\Checkmark&&\Checkmark&&&                           & 36.9/42.5  & 30.4/\textbf{36.2}  & 8.8/11.1    & 7.2/14.4  & 50.5/55.5 &        -   &   -        & -         \\
M5&&\Checkmark&\Checkmark&&&                            & 33.9/39.1  & 27.6/33.6  & \textbf{17.1}/\textbf{17.9}   & \textbf{9.6}/\textbf{16.8}  & 52.0/54.8 &    -       &          - &    -       \\
M6&\Checkmark&\Checkmark&\Checkmark&&&                       & 37.1/41.8  & 29.8/35.4  & 16.9/17.0   & 9.2/16.6  & \textbf{50.7}/\textbf{55.6} & 5.6/10.2  & 18.5/20.5 & 39.8/45.6 \\
M7&\Checkmark&\Checkmark&\Checkmark&\Checkmark&\Checkmark&\Checkmark         & 34.5/40.7  & 28.5/34.3  & 16.9/17.6   & 8.2/15.9  & 48.2/54.0 & \textbf{11.7}/\textbf{18.5} & \textbf{24.6}/\textbf{27.9} & \textbf{41.2}/\textbf{52.9} \\
\bottomrule
\end{tabular}}
\end{table*}
\subsubsection{Image-Language Benchmarks}
 We evaluate VALOR$\rm _{L}$ on three image-language benchmarks including text-to-image retrieval, image captioning and VQA. As presented results in Table \ref{sota_image}, VALOR$\rm _{L}$ achieves decent performances on three benchmarks. Specifically,  VALOR$\rm _{L}$ achieves comparable performance with FILIP\cite{yao2021filip} on COCO retrieval benchmark. On COCO caption benchmark,  VALOR$\rm _{L}$ outperforms GIT and achieves comparable results with GIT2 model, with much less paramters and data. On VQAv2 benchmark, VALOR$\rm _{L}$ outperforms similar scaling GIT$\rm _{L}$ with big margins and achieves comparable performances with larger GIT model.


\subsection{Ablation Study}

\subsubsection{Vision-Audio-Language Cross-Modality Learning}
 We first conduct 
  experiments to show the necessity of  vision-audio-language cross-modality learning. Specifically, we train models on 9 benchmarks of 6 tasks with different input modalities, in both w and w/o VALOR pretraining settings. As Figure \ref{fig:audio_eff} shows, compared to using single vision or audio modality, utilizing both modalities can get consistent improvements for  9 benchmarks, under both w or w/o pretraining settings. These results prove that combining two modalities indeed help model understand videos universally. In addition, vision-audio-language pretraining can further enhance model's tri-modality inference capabilities.  Audio modality is  more functional for audiovisual-language tasks than vision-language tasks. For example, introduction of audio modality can improve AVR(VALOR) and VR(MSRVTT) performance by 26.0\% and 14.9\%, respectively. This is because audiovisual-language tasks are directly related to audio, unlike vision-language tasks using audio as an auxiliary modality.

\begin{figure*}[t]
\centering
\includegraphics[width=1.0\linewidth]{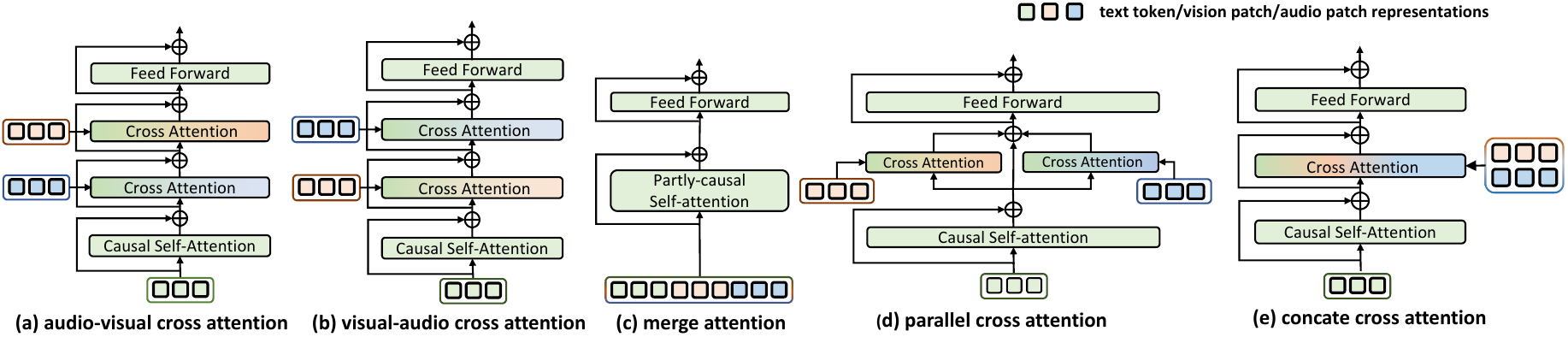}
\caption{Illustrations of variants with different attention mechanisms in multimodal decoder used for MGC task.}
\label{fig:MGC_ablation}
\end{figure*}
\subsubsection{Modality Grouping Strategy}
As introduced in section \ref{sec4.2}, we use modality grouping strategy in both MGA and MGC tasks during pretraining, which aims at enhancing model's generalization capabilities towards tasks with different  modalities input. To demonstrate its effectiveness, we pretrain models  with different modality groups, and evaluate on multiple benchmarks. Taking MGA as an example, from  results presented in Table 
\ref{MGAA} we can get following conclusions. \uppercase\expandafter{\romannumeral1})
Using T-V (M1) or T-A groups (M2) separately for MGA pretraining can get 
relatively good results on corresponding T-V or T-A  benchmarks, but low performance  on T-AV  benchmark.  By contrast, pretrained with  T-AV  group (M3) can achieve 
better performance on  T-AV benchmark, but its performances on T-V and T-A  benchmarks  are separately weaker than M1 and M2, due to the inconsistency between tri-modality pretraining and dual-modality  adapting. 
\uppercase\expandafter{\romannumeral2}) Based on M3, additionally introducing  T-V or T-A group can help enhance  corresponding benchmark performance, but decreases the other. For example, M4 achieves better performances on  T-V benchmarks  than  M3, and are comparable to M1,  but its  performance on T-A benchmark drops harder compared to M3. \uppercase\expandafter{\romannumeral3}) The model trained with T-AV+T-V+T-A groups (M6) can achieves decent performances on  T-AV, T-V and T-A benchmarks, under both zero-shot and finetuning settings. \uppercase\expandafter{\romannumeral4}) Based on M6, further introducing V-A, A-TV
and V-TA groups can improve  performances on corresponding benchmarks, but will also result in evident performance drops for main-stream T-AV, T-V and T-A benchmarks. To this end we  choose M6 as default settings. Besides MGA, modality grouping strategy also functions at MGC task, and similar conclusions can be observed from Table \ref{MGC}.

\begin{table}
\centering
\caption{Downstream  performances of models pretrained on VALOR-1M with  different modality groups in MGC task. Only $ L_{MGC}$ is used. Zero-shot CIDEr / finetune CIDEr are reported.}
\label{MGC}
\scalebox{0.9}{
\begin{tabular}{ccccccccc}
\toprule
\multicolumn{3}{c}{Modality Groups} & \multirow{2}{*}{AVC (VALOR-32K)} & \multirow{2}{*}{VC (MSVD)}   & \multirow{2}{*}{AC (Clotho)} \\
\cmidrule (lr){1-3}
 T-V&T-A&T-AV \\

\midrule
\Checkmark&&               & 31.5/47.5        & 15.4/\textbf{122.6} & 1.6/28.2   \\
&\Checkmark&                & 20.7/39.2        & 0.8/95.0   & 7.6/39.2   \\
&&\Checkmark              & \textbf{43.1}/\textbf{50.0}        & 12.5/119.9 & 10.5/38.1  \\
\Checkmark&&\Checkmark            & 41.6/49.9        & 14.7/121.7 & 6.6/36.2   \\
&\Checkmark&\Checkmark            & 42.7/49.6        & 11.5/116.7 & \textbf{8.1}/40.1   \\
\Checkmark&\Checkmark&\Checkmark          & 41.4/49.8        & \textbf{15.6}/122.2 & 8.0/\textbf{40.4}  \\
\bottomrule
\end{tabular}}
\end{table}

\begin{table}[]
\caption{Downstream  performances of models pretrained on VALOR-1M with  different audiovisual fusion methods for MGA  task.  Models are pretrained on VALOR-1M with $ L_{MGA (T-AV)}$ only. MSR is  short for  MSRVTT dataset. Bold lines are default setting.}
\label{AVR_ablation}
\scalebox{0.9}{
\begin{tabular}{lllll}
\toprule
\multirow{2}{*}{Method}   & \multicolumn{2}{c}{AVR (VALOR-32K)} & \multicolumn{2}{c}{VR (MSR)} \\
\cmidrule (lr){2-3} \cmidrule (lr){4-5}
        & R@1          & R@5          & R@1          & R@5          \\
\midrule
Coarse +  score fusion                              & 53.4         & 81.3         & 35.0         & 66.0         \\
Coarse + feature fusion                             & 54.9         & 81.5         & 34.9         & 66.0         \\
Fine + score fusion                                 & 54.5         & 82.3         & 35.7         & \textbf{67.0}         \\
Fine + feature fusion                               & 54.9         & 81.7         & 36.7         & 65.9         \\
\textbf{Fine + feature fusion + weighted avg}                & \textbf{55.4}         & \textbf{82.7}         & \textbf{36.8}         & 66.5         \\
\bottomrule
\end{tabular}}
\end{table}

\subsubsection{Audiovisual Fusion}
\textbf{Audiovisual Fusion in MGA (T-AV).} MGA task with T-AV group aims at building fine-grained alignment between  language and  the fusion of audio and vision. We compare it to coarse-grained alignment counterpart, in which  global representation of a whole sentence is aligned with the fusion of global representations of whole video and audio. In addition, we compare two audiovisual fusion methods, including feature fusion and score fusion. Feature fusion fuse audio and visual features first  (concatenate them along hidden dimension for coarse-grained alignment or  along sequence dimension for fine-grained alignment) before computing similarity with texts, while score fusion independently compute text-video and text-audio similarity scores, and then add them as total scores. From  results shown in Table \ref{AVR_ablation}, we find that  fine-grained alignment combined with feature fusion achieves best results on both VALOR-32K and MSRVTT datasets, among all four combinations.  Using weighted average introduced in Section 4.2 can further improve  performance consistently on  two benchmarks.

\begin{table}[t]
\caption{Downstream  performances of models pretrained on VALOR-1M with  different E different audiovisual fusion methods for MGC task.  Models are pretrained on VALOR-1M  with $ L_{MGC (T-AV)}$ only. CIDEr metric is reported for captioning task.Bold lines are default setting.}
\label{AVC_ablation}
\centering
\scalebox{0.87}{
\begin{tabular}{lccc}
\toprule

Method & AVC (VALOR-32K) & VC (MSR) & VQA (MSR) \\ 

\midrule
Merge   attention                            & 48.2                 & 56.0       & 43.4    \\
Audio-visual cross   attention                 & 49.6               & 57.0    &   43.7    \\
Visual-audio cross   attention                 & 49.6               & 57.4    &   43.6    \\
Parallel cross      attention               & 49.5                  & 57.4       & 43.8   \\
\textbf{Concate cross       attention}                  & \textbf{50.0}                  & \textbf{59.0}   &  \textbf{44.1}      \\
 
\bottomrule
\end{tabular}}
\end{table}

\begin{table}[]
\caption{Downstream  performances of models pretrained on VALOR-1M with  different MGA and MGC tasks combination settings. R@1 and CIDEr is reported for retrieval and captioning, respectively.}

\label{combine_tasks}
\centering
\scalebox{0.85}{
\begin{tabular}{ccccc}
\toprule
Share weights & $\alpha$ & AVR (VALOR-32K)        & AVC (VALOR-32K) & VQA (MSR) \\
\midrule
\ding{55}            & 1           & 53.8 & 50.3   & 44.1   \\
\Checkmark            & 1           & 54.7& 50.0   & 44.1  \\
\Checkmark             & 0.5       & 53.0 & \textbf{50.8}  & \textbf{44.1}  \\
\Checkmark             & 1.5       & 54.9 & 50.4   & 44.0   \\
\Checkmark             & 3.0       & \textbf{55.4} & 49.9   & 43.8  \\
\bottomrule
\end{tabular}}
\end{table}

\begin{table*}[t]
\caption{Downstream  performances of models with different architectures. Both vision and audio are used for pretraining and finetuning. All models use  AST$_{\rm B}$ as audio encoder.}
\label{encoder}
\centering
\scalebox{1.0}{
\begin{tabular}{lllcccccc}

\toprule
Vision encoder & Text encoder  &\#Example &AVR (VALOR-32K)  & VR (MSRTT)& VR (DiDeMo) & VR (LSMDC) \\
\midrule
Video Swin$_{\rm B}$ & BERT$_{\rm B}$ & 1M & 56.8/83.1/89.9 &39.3/69.0/80.4&41.1/73.1/81.8& 19.8/41.3/52.5\\
CLIP$_{\rm B}$ & BERT$_{\rm B}$ & 1M &61.8/86.0/92.3 &45.0/74.1/84.4&46.3/75.6/84.2   & 22.7/45.4/56.1\\
CLIP$_{\rm L}$ & BERT$_{\rm B}$ & 1M & 64.3/87.3/92.8 &47.7/76.1/84.6&47.3/77.9/85.1&27.8/51.4/60.9 \\

CLIP$_{\rm L}$ & CLIP$_{\rm L}$ & 1M &59.3/84.4/91.0 &50.6/79.0/87.8&53.4/80.2/88.3& 29.2/51.6/58.8\\

CLIP$_{\rm L}$ & BERT$_{\rm L}$ & 1M & 66.5/88.1/93.4 &46.1/76.7/85.4&49.0/77.4/84.9&29.2/50.7/60.5 \\

CLIP$_{\rm L}$ & BERT$_{\rm B}$ & 33.5M              & \textbf{73.2}/\textbf{91.6}/\textbf{95.4}   & 54.4/79.8/87.6    & \textbf{57.6}/\textbf{83.3}/\textbf{88.8}    & 31.8/\textbf{52.8}/62.4  &  \\
CLIP$_{\rm L}$ & CLIP$_{\rm L}$ & 33.5M                    & 67.8/89.5/94.0    & \textbf{55.3}/\textbf{80.5}/\textbf{88.1}    & 57.1/82.9/88.6    &\textbf{ 32.6}/52.6/\textbf{62.7} \\
\bottomrule
\end{tabular}}
\end{table*}

\begin{figure*}[t]
\centering
\includegraphics[width=1.0\linewidth]{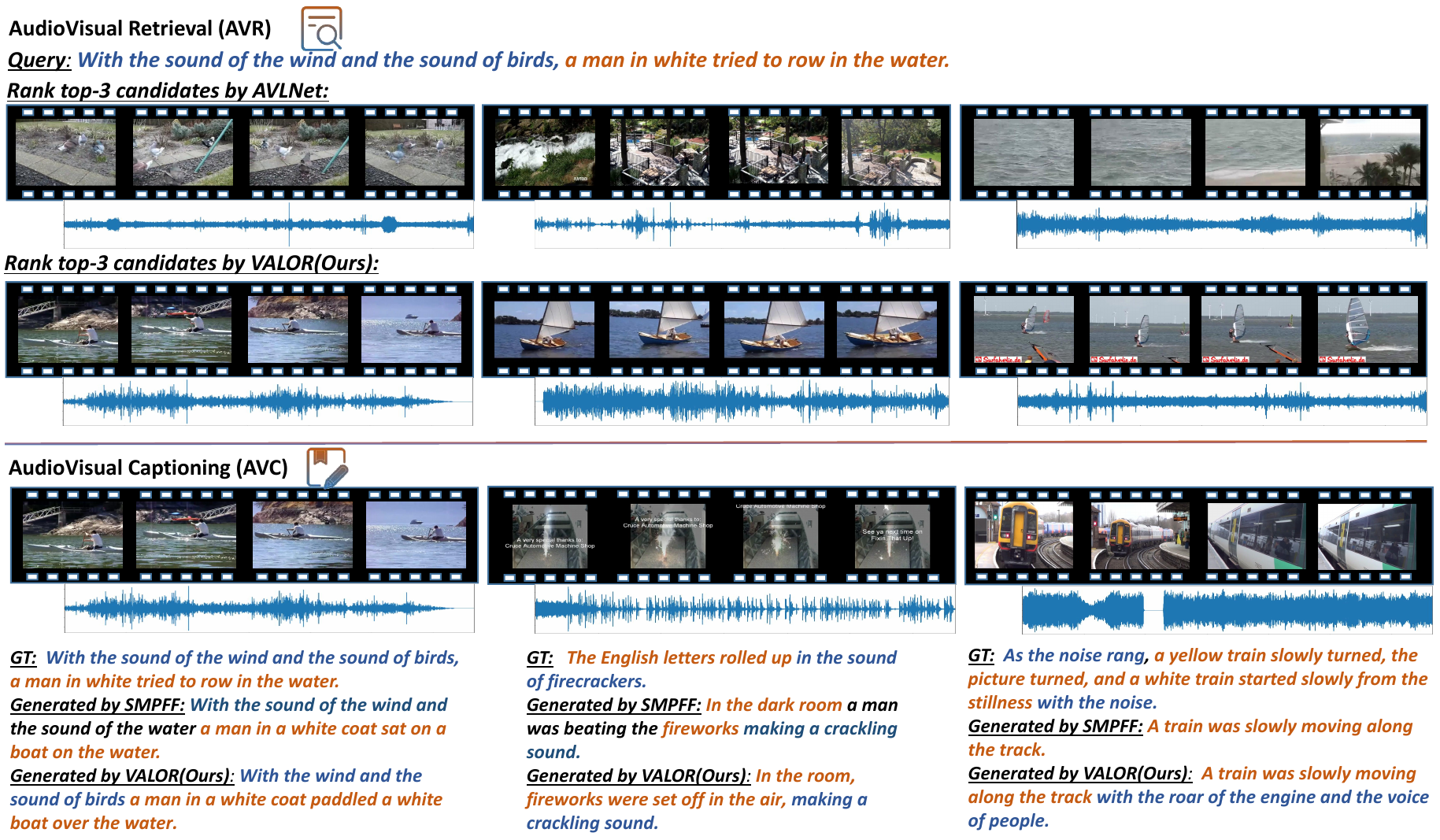}
\caption{Visualization of prediction results of different models  on VALOR-32K  benchmarks. Click the bottons to play the audio.}
\label{visual_model}
\end{figure*}

\textbf{Audiovisual Fusion in MGC (T-AV).} MGC  task with T-AV group demands model to predict  masked tokens with both visual and audio feature as conditions. We make comparison among five variants with different attention mechanisms  in multimodal decoder,   which is illustrated in Figure \ref{fig:MGC_ablation}.  Specifically, audio-visual cross-attention (a) and visual-audio  cross attention (b) variants introduce two cross-attention layers and attends to two modalities in order. Merge attention  (c) directly concatenates tri-modality features and use part-causal mask for  self-attention layers, enabling vision and audio  can fully attend to each other, while preventing information leakage for text. Concatenate cross attention  (e) introduces one cross-attention layer and  attends to the concatenation of two modalities. Parallel cross attention  (d) use independent parameters for two modalities instead of sharing  weights as  (e). From the results shown in Table \ref{AVC_ablation}, we can find  concatenate cross attention mechanism achieves best results consistently on AVC, VC and VQA tasks.

\begin{table}[t]
\caption{Downstream  performances of models pretrained on different  datasets. Models are trained with same iterations on each dataset for fair comparison.}
\centering
\label{Datasets}
\scalebox{0.9}{
\begin{tabular}{lccc}
\toprule
Dataset                   & VR (MSRVTT)        & VC (MSRVTT) & VQA (MSRVTT) \\
\midrule
HD\_VILA\_10M              & 30.4 & 55.4    & 43.2  \\
WebVid-2.5M                & 35.7 & 58.8   & 43.8  \\
CC3M                       & 36.3 & 58.8   & 44.0   \\

VALOR-1M                  & \textbf{39.5} & \textbf{60.7}   & \textbf{44.5}  \\
\bottomrule
\end{tabular}}
\end{table}

\textbf{Combination of MGA and MGC.} When models are trained together with MGA and MGC tasks, we share the common parameters of text encoder and decoder, and  a hyperpameter $\alpha$ is used  to balance two losses. As shown in Table \ref{combine_tasks}, training two tasks together get performance drop on  AVR  (53.8 vs 55.6) and improvement on AVC  (50.3 vs 49.6) tasks, compared to the results of separate training  in Table \ref{MGAA} and Table \ref{MGC}. Parameter sharing can improve 0.9 points on AVR task,  with only slight  decrease on AVC task. Using a larger $\alpha$ will make model focus more on MGA task and get higher AVR performance, but AVC performance gets lower. Video QA is relatively not sensitive to different settings. To the end, we choose parameter sharing and set $\alpha=1.5$, for its decent performances on multiple tasks.

\subsubsection{Effect of VALOR-1M Dataset}
In Table \ref{Datasets}, we  make comparison between VALOR-1M and other public video-language pretraining datasets including WebVid-2.5M, CC3M and HD\_VILA\_10M. The model pretrained on  VALOR-1M use both vision and audio modalities, while models pretrained on other datasets only use vision modality. All models are finetuned on MSRVTT dataset with both visual and audio modalities. From the results we can find that the model pretrained on VALOR-1M surpasses other models  on all three  benchmarks with evident margins, thanks to the high-quality audiovisual captions in VALOR-1M.

\subsubsection{Model Architecture Choices}
We make comparisons about different model architectures in Table \ref{encoder}. As the results shown, when BERT$_{\rm B}$ is chosen as text encoder, using more powerful vision encoder gives more improvement on all four retrieval benchmarks. In addition, we make comparison of text encoders when  CLIP$_{\rm L}$  is chosen as vision encoder. If using BERT$_{\rm B}$ as text encoder,   three single-modality encoders are not aligned at the start of MGA learning. By contrast, if using CLIP$_{\rm L}$ as text encoder, vision and text have already been aligned in advance before MGA learning.  As the results  shown, when pretraining data is limited to VALOR-1M, CLIP$_{\rm L}$ evidently outperforms BERT$_{\rm B}$ on 3 VR benchmarks, due to the benifit of large scale CLIP contrastive pretraining. However, performance of CLIP$_{\rm L}$ on  AVR benchmark   fall behinds BERT$_{\rm B}$, we assume that pre-aligning vision and language makes model tend to ignore the learning of   audiovisual-language correlation, cause simply utilizing vision information can result in  a small loss at the start of pretraining. When utilizing more training data (33.5M), the disadvantage of BERT$_{\rm B}$ on VR benchmarks gets largely relieved while its advantage on AVR benchmark remains. In addition, we also tried to scale up text encoder from BERT$_{\rm B}$ to BERT$_{\rm L}$,  and observed performance gains on three benchmarks except for MSRVTT. In the end, we choose CLIP$_{\rm L}$ and BERT$_{\rm B}$ as VALOR$_{\rm L}$'s deafult design for its high effectiveness and efficiency.

\subsection{Visualizations}

In Figure \ref{visual_model}, we make qualitative comparisons on VALOR-32K benchmarks between 
 VALOR$\rm _{B}$ and task-specific methods including AVLNet\cite{rouditchenko2020avlnet} and SMPFF\cite{chen2021mm21}, both of which take vision and audio as inputs.  From the visualization results we can find that compared to AVLNet, VALOR can accurately rank video candidates  given audiovisual text query and successfully retrieve the groundtruth one. Compared to  the rank\#1 video, the rank\#2 video also shows a man in white on water, and is accompanied with wind sound, but  the bird sound is missing. The rank\#3 video is more conflicted to the query, from both vision and audio perspective. With regards to AVC task, SMPFF either recognizes sound wrongly (in example 1), or describes vision wrongly (in example 2), or totally ignores audio information (in example 3). By contrast, VALOR can comprehensively recognize both visual and audio concepts, and generate accurate descriptions for all three examples.

\section{Conclusion}
This paper proposed a unified vision-audio-language cross-modality pretraining model VALOR, which 
 models tri-modality understanding and generation through two designed pretraing tasks including Multimodal Grouping Alignment and Multimodal Grouping Captioning. Extensive experiments have been conducted to demonstrate that VALOR possesses good versatility and  scalability. The first strong correlated vision-audio-language dataset VALOR-1M   is proposed to promote tri-modality pretraining research and  VALOR-32K is proposed for audiovisual-language retrieval and captioning benchmarking. Trained on VALOR-1M and other public vision-language datasets, VALOR achieves series of new state-of-the-art performances on downstream  vision/audio/audiovisual retrieval, captioning and question answering tasks. In future, we plan to increase the scaling of VALOR-1M dataset via unspervised methods like generating and filtering pesudo audiovisual captions. In addition, we also plan to additionally introduce   vision and audio generation modeling into current VALOR framework.

{\small
    \bibliographystyle{IEEEtran}
    \bibliography{VALOR}

\begin{thebibliography}{100}
\providecommand{\url}[1]{#1}
\csname url@samestyle\endcsname
\providecommand{\newblock}{\relax}
\providecommand{\bibinfo}[2]{#2}
\providecommand{\BIBentrySTDinterwordspacing}{\spaceskip=0pt\relax}
\providecommand{\BIBentryALTinterwordstretchfactor}{4}
\providecommand{\BIBentryALTinterwordspacing}{\spaceskip=\fontdimen2\font plus
\BIBentryALTinterwordstretchfactor\fontdimen3\font minus
  \fontdimen4\font\relax}
\providecommand{\BIBforeignlanguage}[2]{{%
\expandafter\ifx\csname l@#1\endcsname\relax
\typeout{** WARNING: IEEEtran.bst: No hyphenation pattern has been}%
\typeout{** loaded for the language `#1'. Using the pattern for}%
\typeout{** the default language instead.}%
\else
\language=\csname l@#1\endcsname
\fi
#2}}
\providecommand{\BIBdecl}{\relax}
\BIBdecl

\bibitem{kim2018bilinear}
J.-H. Kim, J.~Jun, and B.-T. Zhang, ``Bilinear attention networks,''
  \emph{Advances in neural information processing systems}, vol.~31, 2018.

\bibitem{dong2021dual}
J.~Dong, X.~Li, C.~Xu, X.~Yang, G.~Yang, X.~Wang, and M.~Wang, ``Dual encoding
  for video retrieval by text,'' \emph{IEEE Transactions on Pattern Analysis
  and Machine Intelligence}, 2021.

\bibitem{liu2021aligning}
F.~Liu, X.~Wu, C.~You, S.~Ge, Y.~Zou, and X.~Sun, ``Aligning source visual and
  target language domains for unpaired video captioning,'' \emph{IEEE
  Transactions on Pattern Analysis and Machine Intelligence}, 2021.

\bibitem{yan2019fine}
Y.~Yan, N.~Zhuang, B.~Ni, J.~Zhang, M.~Xu, Q.~Zhang, Z.~Zhang, S.~Cheng,
  Q.~Tian, Y.~Xu \emph{et~al.}, ``Fine-grained video captioning via graph-based
  multi-granularity interaction learning,'' \emph{IEEE transactions on pattern
  analysis and machine intelligence}, vol.~44, no.~2, pp. 666--683, 2019.

\bibitem{wang2019controllable}
B.~Wang, L.~Ma, W.~Zhang, W.~Jiang, J.~Wang, and W.~Liu, ``Controllable video
  captioning with pos sequence guidance based on gated fusion network,'' in
  \emph{Proceedings of the IEEE/CVF international conference on computer
  vision}, 2019, pp. 2641--2650.

\bibitem{anderson2018bottom}
P.~Anderson, X.~He, C.~Buehler, D.~Teney, M.~Johnson, S.~Gould, and L.~Zhang,
  ``Bottom-up and top-down attention for image captioning and visual question
  answering,'' in \emph{Proceedings of the IEEE conference on computer vision
  and pattern recognition}, 2018, pp. 6077--6086.

\bibitem{peng2020mra}
L.~Peng, Y.~Yang, Z.~Wang, Z.~Huang, and H.~T. Shen, ``Mra-net: Improving vqa
  via multi-modal relation attention network,'' \emph{IEEE Transactions on
  Pattern Analysis and Machine Intelligence}, vol.~44, no.~1, pp. 318--329,
  2020.

\bibitem{devlin2018bert}
J.~Devlin, M.-W. Chang, K.~Lee, and K.~Toutanova, ``Bert: Pre-training of deep
  bidirectional transformers for language understanding,'' \emph{arXiv preprint
  arXiv:1810.04805}, 2018.

\bibitem{radford2018improving}
A.~Radford, K.~Narasimhan, T.~Salimans, I.~Sutskever \emph{et~al.}, ``Improving
  language understanding by generative pre-training,'' 2018.

\bibitem{brown2020language}
T.~Brown, B.~Mann, N.~Ryder, M.~Subbiah, J.~D. Kaplan, P.~Dhariwal,
  A.~Neelakantan, P.~Shyam, G.~Sastry, A.~Askell \emph{et~al.}, ``Language
  models are few-shot learners,'' \emph{Advances in neural information
  processing systems}, vol.~33, pp. 1877--1901, 2020.

\bibitem{miech2019howto100m}
A.~Miech, D.~Zhukov, J.-B. Alayrac, M.~Tapaswi, I.~Laptev, and J.~Sivic,
  ``Howto100m: Learning a text-video embedding by watching hundred million
  narrated video clips,'' in \emph{Proceedings of the IEEE/CVF International
  Conference on Computer Vision}, 2019, pp. 2630--2640.

\bibitem{xue2022advancing}
H.~Xue, T.~Hang, Y.~Zeng, Y.~Sun, B.~Liu, H.~Yang, J.~Fu, and B.~Guo,
  ``Advancing high-resolution video-language representation with large-scale
  video transcriptions,'' in \emph{Proceedings of the IEEE/CVF Conference on
  Computer Vision and Pattern Recognition}, 2022, pp. 5036--5045.

\bibitem{bain2021frozen}
M.~Bain, A.~Nagrani, G.~Varol, and A.~Zisserman, ``Frozen in time: A joint
  video and image encoder for end-to-end retrieval,'' in \emph{Proceedings of
  the IEEE/CVF International Conference on Computer Vision}, 2021, pp.
  1728--1738.

\bibitem{wang2022git}
J.~Wang, Z.~Yang, X.~Hu, L.~Li, K.~Lin, Z.~Gan, Z.~Liu, C.~Liu, and L.~Wang,
  ``Git: A generative image-to-text transformer for vision and language,''
  \emph{arXiv preprint arXiv:2205.14100}, 2022.

\bibitem{zellers2022merlot}
R.~Zellers, J.~Lu, X.~Lu, Y.~Yu, Y.~Zhao, M.~Salehi, A.~Kusupati, J.~Hessel,
  A.~Farhadi, and Y.~Choi, ``Merlot reserve: Neural script knowledge through
  vision and language and sound,'' in \emph{Proceedings of the IEEE/CVF
  Conference on Computer Vision and Pattern Recognition}, 2022, pp.
  16\,375--16\,387.

\bibitem{miech2020end}
A.~Miech, J.-B. Alayrac, L.~Smaira, I.~Laptev, J.~Sivic, and A.~Zisserman,
  ``End-to-end learning of visual representations from uncurated instructional
  videos,'' in \emph{Proceedings of the IEEE/CVF Conference on Computer Vision
  and Pattern Recognition}, 2020, pp. 9879--9889.

\bibitem{sharma2018conceptual}
P.~Sharma, N.~Ding, S.~Goodman, and R.~Soricut, ``Conceptual captions: A
  cleaned, hypernymed, image alt-text dataset for automatic image captioning,''
  in \emph{Proceedings of the 56th Annual Meeting of the Association for
  Computational Linguistics (Volume 1: Long Papers)}, 2018, pp. 2556--2565.

\bibitem{changpinyo2021conceptual}
S.~Changpinyo, P.~Sharma, N.~Ding, and R.~Soricut, ``Conceptual 12m: Pushing
  web-scale image-text pre-training to recognize long-tail visual concepts,''
  in \emph{Proceedings of the IEEE/CVF Conference on Computer Vision and
  Pattern Recognition}, 2021, pp. 3558--3568.

\bibitem{seo2022end}
P.~H. Seo, A.~Nagrani, A.~Arnab, and C.~Schmid, ``End-to-end generative
  pretraining for multimodal video captioning,'' in \emph{Proceedings of the
  IEEE/CVF Conference on Computer Vision and Pattern Recognition}, 2022, pp.
  17\,959--17\,968.

\bibitem{xue2022clip}
H.~Xue, Y.~Sun, B.~Liu, J.~Fu, R.~Song, H.~Li, and J.~Luo, ``Clip-vip: Adapting
  pre-trained image-text model to video-language representation alignment,''
  \emph{arXiv preprint arXiv:2209.06430}, 2022.

\bibitem{chen2020uniter}
Y.-C. Chen, L.~Li, L.~Yu, A.~El~Kholy, F.~Ahmed, Z.~Gan, Y.~Cheng, and J.~Liu,
  ``Uniter: Universal image-text representation learning,'' in \emph{European
  conference on computer vision}.\hskip 1em plus 0.5em minus 0.4em\relax
  Springer, 2020, pp. 104--120.

\bibitem{lu2019vilbert}
J.~Lu, D.~Batra, D.~Parikh, and S.~Lee, ``Vilbert: Pretraining task-agnostic
  visiolinguistic representations for vision-and-language tasks,''
  \emph{Advances in neural information processing systems}, vol.~32, 2019.

\bibitem{fu2022empirical}
T.-J. Fu, L.~Li, Z.~Gan, K.~Lin, W.~Y. Wang, L.~Wang, and Z.~Liu, ``An
  empirical study of end-to-end video-language transformers with masked visual
  modeling,'' \emph{arXiv preprint arXiv:2209.01540}, 2022.

\bibitem{li2020hero}
L.~Li, Y.-C. Chen, Y.~Cheng, Z.~Gan, L.~Yu, and J.~Liu, ``Hero: Hierarchical
  encoder for video+ language omni-representation pre-training,'' in
  \emph{Proceedings of the 2020 Conference on Empirical Methods in Natural
  Language Processing (EMNLP)}, 2020, pp. 2046--2065.

\bibitem{li2021align}
J.~Li, R.~Selvaraju, A.~Gotmare, S.~Joty, C.~Xiong, and S.~C.~H. Hoi, ``Align
  before fuse: Vision and language representation learning with momentum
  distillation,'' \emph{Advances in neural information processing systems},
  vol.~34, pp. 9694--9705, 2021.

\bibitem{ren2015faster}
S.~Ren, K.~He, R.~Girshick, and J.~Sun, ``Faster r-cnn: Towards real-time
  object detection with region proposal networks,'' \emph{Advances in neural
  information processing systems}, vol.~28, 2015.

\bibitem{dosovitskiy2020image}
A.~Dosovitskiy, L.~Beyer, A.~Kolesnikov, D.~Weissenborn, X.~Zhai,
  T.~Unterthiner, M.~Dehghani, M.~Minderer, G.~Heigold, S.~Gelly \emph{et~al.},
  ``An image is worth 16x16 words: Transformers for image recognition at
  scale,'' \emph{arXiv preprint arXiv:2010.11929}, 2020.

\bibitem{liu2021swin}
Z.~Liu, Y.~Lin, Y.~Cao, H.~Hu, Y.~Wei, Z.~Zhang, S.~Lin, and B.~Guo, ``Swin
  transformer: Hierarchical vision transformer using shifted windows,'' in
  \emph{Proceedings of the IEEE/CVF International Conference on Computer
  Vision}, 2021, pp. 10\,012--10\,022.

\bibitem{cho2021unifying}
J.~Cho, J.~Lei, H.~Tan, and M.~Bansal, ``Unifying vision-and-language tasks via
  text generation,'' in \emph{International Conference on Machine
  Learning}.\hskip 1em plus 0.5em minus 0.4em\relax PMLR, 2021, pp. 1931--1942.

\bibitem{yang2022unitab}
Z.~Yang, Z.~Gan, J.~Wang, X.~Hu, F.~Ahmed, Z.~Liu, Y.~Lu, and L.~Wang,
  ``Unitab: Unifying text and box outputs for grounded vision-language
  modeling,'' in \emph{European Conference on Computer Vision}.\hskip 1em plus
  0.5em minus 0.4em\relax Springer, 2022, pp. 521--539.

\bibitem{wang2022ofa}
P.~Wang, A.~Yang, R.~Men, J.~Lin, S.~Bai, Z.~Li, J.~Ma, C.~Zhou, J.~Zhou, and
  H.~Yang, ``Ofa: Unifying architectures, tasks, and modalities through a
  simple sequence-to-sequence learning framework,'' in \emph{International
  Conference on Machine Learning}.\hskip 1em plus 0.5em minus 0.4em\relax PMLR,
  2022, pp. 23\,318--23\,340.

\bibitem{lu2022unified}
J.~Lu, C.~Clark, R.~Zellers, R.~Mottaghi, and A.~Kembhavi, ``Unified-io: A
  unified model for vision, language, and multi-modal tasks,'' \emph{arXiv
  preprint arXiv:2206.08916}, 2022.

\bibitem{Chen2021Pix2seqAL}
T.~Chen, S.~Saxena, L.~Li, D.~J. Fleet, and G.~rey E.~Hinton, ``Pix2seq: A
  language modeling framework for object detection,'' \emph{ArXiv}, vol.
  abs/2109.10852, 2021.

\bibitem{van2017neural}
A.~Van Den~Oord, O.~Vinyals \emph{et~al.}, ``Neural discrete representation
  learning,'' \emph{Advances in neural information processing systems},
  vol.~30, 2017.

\bibitem{zhu2022uni}
X.~Zhu, J.~Zhu, H.~Li, X.~Wu, H.~Li, X.~Wang, and J.~Dai, ``Uni-perceiver:
  Pre-training unified architecture for generic perception for zero-shot and
  few-shot tasks,'' in \emph{Proceedings of the IEEE/CVF Conference on Computer
  Vision and Pattern Recognition}, 2022, pp. 16\,804--16\,815.

\bibitem{li2022lavender}
L.~Li, Z.~Gan, K.~Lin, C.-C. Lin, Z.~Liu, C.~Liu, and L.~Wang, ``Lavender:
  Unifying video-language understanding as masked language modeling,''
  \emph{arXiv preprint arXiv:2206.07160}, 2022.

\bibitem{radford2021learning}
A.~Radford, J.~W. Kim, C.~Hallacy, A.~Ramesh, G.~Goh, S.~Agarwal, G.~Sastry,
  A.~Askell, P.~Mishkin, J.~Clark \emph{et~al.}, ``Learning transferable visual
  models from natural language supervision,'' in \emph{International Conference
  on Machine Learning}.\hskip 1em plus 0.5em minus 0.4em\relax PMLR, 2021, pp.
  8748--8763.

\bibitem{jia2021scaling}
C.~Jia, Y.~Yang, Y.~Xia, Y.-T. Chen, Z.~Parekh, H.~Pham, Q.~Le, Y.-H. Sung,
  Z.~Li, and T.~Duerig, ``Scaling up visual and vision-language representation
  learning with noisy text supervision,'' in \emph{International Conference on
  Machine Learning}.\hskip 1em plus 0.5em minus 0.4em\relax PMLR, 2021, pp.
  4904--4916.

\bibitem{yuan2021florence}
L.~Yuan, D.~Chen, Y.-L. Chen, N.~Codella, X.~Dai, J.~Gao, H.~Hu, X.~Huang,
  B.~Li, C.~Li \emph{et~al.}, ``Florence: A new foundation model for computer
  vision,'' \emph{arXiv preprint arXiv:2111.11432}, 2021.

\bibitem{pham2021combined}
H.~Pham, Z.~Dai, G.~Ghiasi, K.~Kawaguchi, H.~Liu, A.~W. Yu, J.~Yu, Y.-T. Chen,
  M.-T. Luong, Y.~Wu \emph{et~al.}, ``Combined scaling for open-vocabulary
  image classification,'' \emph{arXiv preprint arXiv: 2111.10050}, 2021.

\bibitem{wang2021simvlm}
Z.~Wang, J.~Yu, A.~W. Yu, Z.~Dai, Y.~Tsvetkov, and Y.~Cao, ``Simvlm: Simple
  visual language model pretraining with weak supervision,'' \emph{arXiv
  preprint arXiv:2108.10904}, 2021.

\bibitem{alayrac2022flamingo}
J.-B. Alayrac, J.~Donahue, P.~Luc, A.~Miech, I.~Barr, Y.~Hasson, K.~Lenc,
  A.~Mensch, K.~Millican, M.~Reynolds \emph{et~al.}, ``Flamingo: a visual
  language model for few-shot learning,'' \emph{arXiv preprint
  arXiv:2204.14198}, 2022.

\bibitem{chen2022pali}
X.~Chen, X.~Wang, S.~Changpinyo, A.~Piergiovanni, P.~Padlewski, D.~Salz,
  S.~Goodman, A.~Grycner, B.~Mustafa, L.~Beyer \emph{et~al.}, ``Pali: A
  jointly-scaled multilingual language-image model,'' \emph{arXiv preprint
  arXiv:2209.06794}, 2022.

\bibitem{yu2022coca}
J.~Yu, Z.~Wang, V.~Vasudevan, L.~Yeung, M.~Seyedhosseini, and Y.~Wu, ``Coca:
  Contrastive captioners are image-text foundation models,'' \emph{arXiv
  preprint arXiv:2205.01917}, 2022.

\bibitem{gabeur2020multi}
V.~Gabeur, C.~Sun, K.~Alahari, and C.~Schmid, ``Multi-modal transformer for
  video retrieval,'' in \emph{European Conference on Computer Vision}.\hskip
  1em plus 0.5em minus 0.4em\relax Springer, 2020, pp. 214--229.

\bibitem{chen2021mm21}
S.~Chen, X.~Zhu, D.~Hao, W.~Liu, J.~Liu, Z.~Zhao, L.~Guo, and J.~Liu, ``Mm21
  pre-training for video understanding challenge: Video captioning with
  pretraining techniques,'' in \emph{Proceedings of the 29th ACM International
  Conference on Multimedia}, 2021, pp. 4853--4857.

\bibitem{luo2020univl}
H.~Luo, L.~Ji, B.~Shi, H.~Huang, N.~Duan, T.~Li, J.~Li, T.~Bharti, and M.~Zhou,
  ``Univl: A unified video and language pre-training model for multimodal
  understanding and generation,'' \emph{arXiv preprint arXiv:2002.06353}, 2020.

\bibitem{xu2021vlm}
H.~Xu, G.~Ghosh, P.-Y. Huang, P.~Arora, M.~Aminzadeh, C.~Feichtenhofer,
  F.~Metze, and L.~Zettlemoyer, ``Vlm: Task-agnostic video-language model
  pre-training for video understanding,'' \emph{arXiv preprint
  arXiv:2105.09996}, 2021.

\bibitem{li2021value}
L.~Li, J.~Lei, Z.~Gan, L.~Yu, Y.-C. Chen, R.~Pillai, Y.~Cheng, L.~Zhou, X.~E.
  Wang, W.~Y. Wang \emph{et~al.}, ``Value: A multi-task benchmark for
  video-and-language understanding evaluation,'' \emph{arXiv preprint
  arXiv:2106.04632}, 2021.

\bibitem{rouditchenko2020avlnet}
A.~Rouditchenko, A.~Boggust, D.~Harwath, B.~Chen, D.~Joshi, S.~Thomas,
  K.~Audhkhasi, H.~Kuehne, R.~Panda, R.~Feris \emph{et~al.}, ``Avlnet: Learning
  audio-visual language representations from instructional videos,''
  \emph{arXiv preprint arXiv:2006.09199}, 2020.

\bibitem{chen2021multimodal}
B.~Chen, A.~Rouditchenko, K.~Duarte, H.~Kuehne, S.~Thomas, A.~Boggust,
  R.~Panda, B.~Kingsbury, R.~Feris, D.~Harwath \emph{et~al.}, ``Multimodal
  clustering networks for self-supervised learning from unlabeled videos,'' in
  \emph{Proceedings of the IEEE/CVF International Conference on Computer
  Vision}, 2021, pp. 8012--8021.

\bibitem{akbari2021vatt}
H.~Akbari, L.~Yuan, R.~Qian, W.-H. Chuang, S.-F. Chang, Y.~Cui, and B.~Gong,
  ``Vatt: Transformers for multimodal self-supervised learning from raw video,
  audio and text,'' \emph{Advances in Neural Information Processing Systems},
  vol.~34, pp. 24\,206--24\,221, 2021.

\bibitem{yang2022code}
Z.~Yang, Y.~Fang, C.~Zhu, R.~Pryzant, D.~Chen, Y.~Shi, Y.~Xu, Y.~Qian, M.~Gao,
  Y.-L. Chen \emph{et~al.}, ``i-code: An integrative and composable multimodal
  learning framework,'' \emph{arXiv preprint arXiv:2205.01818}, 2022.

\bibitem{chen2011collecting}
D.~Chen and W.~B. Dolan, ``Collecting highly parallel data for paraphrase
  evaluation,'' in \emph{Proceedings of the 49th annual meeting of the
  association for computational linguistics: human language technologies},
  2011, pp. 190--200.

\bibitem{xu2016msr}
J.~Xu, T.~Mei, T.~Yao, and Y.~Rui, ``Msr-vtt: A large video description dataset
  for bridging video and language,'' in \emph{Proceedings of the IEEE
  conference on computer vision and pattern recognition}, 2016, pp. 5288--5296.

\bibitem{wang2019vatex}
X.~Wang, J.~Wu, J.~Chen, L.~Li, Y.-F. Wang, and W.~Y. Wang, ``Vatex: A
  large-scale, high-quality multilingual dataset for video-and-language
  research,'' in \emph{Proceedings of the IEEE/CVF International Conference on
  Computer Vision}, 2019, pp. 4581--4591.

\bibitem{zhou2018towards}
L.~Zhou, C.~Xu, and J.~J. Corso, ``Towards automatic learning of procedures
  from web instructional videos,'' in \emph{Thirty-Second AAAI Conference on
  Artificial Intelligence}, 2018.

\bibitem{anne2017localizing}
L.~Anne~Hendricks, O.~Wang, E.~Shechtman, J.~Sivic, T.~Darrell, and B.~Russell,
  ``Localizing moments in video with natural language,'' in \emph{Proceedings
  of the IEEE international conference on computer vision}, 2017, pp.
  5803--5812.

\bibitem{krishna2017dense}
R.~Krishna, K.~Hata, F.~Ren, L.~Fei-Fei, and J.~Carlos~Niebles,
  ``Dense-captioning events in videos,'' in \emph{Proceedings of the IEEE
  international conference on computer vision}, 2017, pp. 706--715.

\bibitem{rohrbach2017movie}
A.~Rohrbach, A.~Torabi, M.~Rohrbach, N.~Tandon, C.~Pal, H.~Larochelle,
  A.~Courville, and B.~Schiele, ``Movie description,'' \emph{International
  Journal of Computer Vision}, vol. 123, no.~1, pp. 94--120, 2017.

\bibitem{drossos2020clotho}
K.~Drossos, S.~Lipping, and T.~Virtanen, ``Clotho: An audio captioning
  dataset,'' in \emph{ICASSP 2020-2020 IEEE International Conference on
  Acoustics, Speech and Signal Processing (ICASSP)}.\hskip 1em plus 0.5em minus
  0.4em\relax IEEE, 2020, pp. 736--740.

\bibitem{kim2019audiocaps}
C.~D. Kim, B.~Kim, H.~Lee, and G.~Kim, ``Audiocaps: Generating captions for
  audios in the wild,'' in \emph{Proceedings of the 2019 Conference of the
  North American Chapter of the Association for Computational Linguistics:
  Human Language Technologies, Volume 1 (Long and Short Papers)}, 2019, pp.
  119--132.

\bibitem{yun2021pano}
H.~Yun, Y.~Yu, W.~Yang, K.~Lee, and G.~Kim, ``Pano-avqa: Grounded audio-visual
  question answering on 360deg videos,'' in \emph{Proceedings of the IEEE/CVF
  International Conference on Computer Vision}, 2021, pp. 2031--2041.

\bibitem{li2022learning}
G.~Li, Y.~Wei, Y.~Tian, C.~Xu, J.-R. Wen, and D.~Hu, ``Learning to answer
  questions in dynamic audio-visual scenarios,'' in \emph{Proceedings of the
  IEEE/CVF Conference on Computer Vision and Pattern Recognition}, 2022, pp.
  19\,108--19\,118.

\bibitem{yang2022avqa}
P.~Yang, X.~Wang, X.~Duan, H.~Chen, R.~Hou, C.~Jin, and W.~Zhu, ``Avqa: A
  dataset for audio-visual question answering on videos,'' in \emph{Proceedings
  of the 30th ACM International Conference on Multimedia}, 2022, pp.
  3480--3491.

\bibitem{gemmeke2017audio}
J.~F. Gemmeke, D.~P. Ellis, D.~Freedman, A.~Jansen, W.~Lawrence, R.~C. Moore,
  M.~Plakal, and M.~Ritter, ``Audio set: An ontology and human-labeled dataset
  for audio events,'' in \emph{2017 IEEE international conference on acoustics,
  speech and signal processing (ICASSP)}.\hskip 1em plus 0.5em minus
  0.4em\relax IEEE, 2017, pp. 776--780.

\bibitem{liu2022video}
Z.~Liu, J.~Ning, Y.~Cao, Y.~Wei, Z.~Zhang, S.~Lin, and H.~Hu, ``Video swin
  transformer,'' in \emph{Proceedings of the IEEE/CVF Conference on Computer
  Vision and Pattern Recognition}, 2022, pp. 3202--3211.

\bibitem{gong2021ast}
Y.~Gong, Y.-A. Chung, and J.~Glass, ``Ast: Audio spectrogram transformer,''
  \emph{arXiv preprint arXiv:2104.01778}, 2021.

\bibitem{gong2022ssast}
Y.~Gong, C.-I. Lai, Y.-A. Chung, and J.~Glass, ``Ssast: Self-supervised audio
  spectrogram transformer,'' in \emph{Proceedings of the AAAI Conference on
  Artificial Intelligence}, vol.~36, no.~10, 2022, pp. 10\,699--10\,709.

\bibitem{cheng2021improving}
X.~Cheng, H.~Lin, X.~Wu, F.~Yang, and D.~Shen, ``Improving video-text retrieval
  by multi-stream corpus alignment and dual softmax loss,'' \emph{arXiv
  preprint arXiv:2109.04290}, 2021.

\bibitem{lei2021less}
J.~Lei, L.~Li, L.~Zhou, Z.~Gan, T.~L. Berg, M.~Bansal, and J.~Liu, ``Less is
  more: Clipbert for video-and-language learning via sparse sampling,'' in
  \emph{Proceedings of the IEEE/CVF Conference on Computer Vision and Pattern
  Recognition}, 2021, pp. 7331--7341.

\bibitem{ge2022bridging}
Y.~Ge, Y.~Ge, X.~Liu, D.~Li, Y.~Shan, X.~Qie, and P.~Luo, ``Bridging video-text
  retrieval with multiple choice questions,'' in \emph{Proceedings of the
  IEEE/CVF Conference on Computer Vision and Pattern Recognition}, 2022, pp.
  16\,167--16\,176.

\bibitem{ge2022miles}
Y.~Ge, Y.~Ge, X.~Liu, A.~J. Wang, J.~Wu, Y.~Shan, X.~Qie, and P.~Luo, ``Miles:
  Visual bert pre-training with injected language semantics for video-text
  retrieval,'' \emph{arXiv preprint arXiv:2204.12408}, 2022.

\bibitem{wang2022object}
J.~Wang, Y.~Ge, G.~Cai, R.~Yan, X.~Lin, Y.~Shan, X.~Qie, and M.~Z. Shou,
  ``Object-aware video-language pre-training for retrieval,'' in
  \emph{Proceedings of the IEEE/CVF Conference on Computer Vision and Pattern
  Recognition}, 2022, pp. 3313--3322.

\bibitem{nagrani2022learning}
A.~Nagrani, P.~H. Seo, B.~Seybold, A.~Hauth, S.~Manen, C.~Sun, and C.~Schmid,
  ``Learning audio-video modalities from image captions,'' \emph{arXiv preprint
  arXiv:2204.00679}, 2022.

\bibitem{sun2022long}
Y.~Sun, H.~Xue, R.~Song, B.~Liu, H.~Yang, and J.~Fu, ``Long-form video-language
  pre-training with multimodal temporal contrastive learning,'' \emph{arXiv
  preprint arXiv:2210.06031}, 2022.

\bibitem{lei2022revealing}
J.~Lei, T.~L. Berg, and M.~Bansal, ``Revealing single frame bias for
  video-and-language learning,'' \emph{arXiv preprint arXiv:2206.03428}, 2022.

\bibitem{yang2021taco}
J.~Yang, Y.~Bisk, and J.~Gao, ``Taco: Token-aware cascade contrastive learning
  for video-text alignment,'' in \emph{Proceedings of the IEEE/CVF
  International Conference on Computer Vision}, 2021, pp. 11\,562--11\,572.

\bibitem{patrick2020support}
M.~Patrick, P.-Y. Huang, Y.~Asano, F.~Metze, A.~Hauptmann, J.~Henriques, and
  A.~Vedaldi, ``Support-set bottlenecks for video-text representation
  learning,'' \emph{arXiv preprint arXiv:2010.02824}, 2020.

\bibitem{gabeur2022masking}
V.~Gabeur, A.~Nagrani, C.~Sun, K.~Alahari, and C.~Schmid, ``Masking modalities
  for cross-modal video retrieval,'' in \emph{Proceedings of the IEEE/CVF
  Winter Conference on Applications of Computer Vision}, 2022, pp. 1766--1775.

\bibitem{wang2022all}
A.~J. Wang, Y.~Ge, R.~Yan, Y.~Ge, X.~Lin, G.~Cai, J.~Wu, Y.~Shan, X.~Qie, and
  M.~Z. Shou, ``All in one: Exploring unified video-language pre-training,''
  \emph{arXiv preprint arXiv:2203.07303}, 2022.

\bibitem{fu2021violet}
T.-J. Fu, L.~Li, Z.~Gan, K.~Lin, W.~Y. Wang, L.~Wang, and Z.~Liu, ``Violet:
  End-to-end video-language transformers with masked visual-token modeling,''
  \emph{arXiv preprint arXiv:2111.12681}, 2021.

\bibitem{luo2022clip4clip}
H.~Luo, L.~Ji, M.~Zhong, Y.~Chen, W.~Lei, N.~Duan, and T.~Li, ``Clip4clip: An
  empirical study of clip for end to end video clip retrieval and captioning,''
  \emph{Neurocomputing}, vol. 508, pp. 293--304, 2022.

\bibitem{liu2022ts2}
Y.~Liu, P.~Xiong, L.~Xu, S.~Cao, and Q.~Jin, ``Ts2-net: Token shift and
  selection transformer for text-video retrieval,'' in \emph{European
  Conference on Computer Vision}.\hskip 1em plus 0.5em minus 0.4em\relax
  Springer, 2022, pp. 319--335.

\bibitem{ma2022x}
Y.~Ma, G.~Xu, X.~Sun, M.~Yan, J.~Zhang, and R.~Ji, ``X-clip: End-to-end
  multi-grained contrastive learning for video-text retrieval,'' in
  \emph{Proceedings of the 30th ACM International Conference on Multimedia},
  2022, pp. 638--647.

\bibitem{lin2022eclipse}
Y.-B. Lin, J.~Lei, M.~Bansal, and G.~Bertasius, ``Eclipse: Efficient long-range
  video retrieval using sight and sound,'' \emph{arXiv preprint
  arXiv:2204.02874}, 2022.

\bibitem{wang2022disentangled}
Q.~Wang, Y.~Zhang, Y.~Zheng, P.~Pan, and X.-S. Hua, ``Disentangled
  representation learning for text-video retrieval,'' \emph{arXiv preprint
  arXiv:2203.07111}, 2022.

\bibitem{min2022hunyuan_tvr}
S.~Min, W.~Kong, R.-C. Tu, D.~Gong, C.~Cai, W.~Zhao, C.~Liu, S.~Zheng, H.~Wang,
  Z.~Li \emph{et~al.}, ``Hunyuan\_tvr for text-video retrivial,'' \emph{arXiv
  preprint arXiv:2204.03382}, 2022.

\bibitem{wang2022internvideo}
Y.~Wang, K.~Li, Y.~Li, Y.~He, B.~Huang, Z.~Zhao, H.~Zhang, J.~Xu, Y.~Liu,
  Z.~Wang \emph{et~al.}, ``Internvideo: General video foundation models via
  generative and discriminative learning,'' \emph{arXiv preprint
  arXiv:2212.03191}, 2022.

\bibitem{lin2014microsoft}
T.-Y. Lin, M.~Maire, S.~Belongie, J.~Hays, P.~Perona, D.~Ramanan,
  P.~Doll{\'a}r, and C.~L. Zitnick, ``Microsoft coco: Common objects in
  context,'' in \emph{European conference on computer vision}.\hskip 1em plus
  0.5em minus 0.4em\relax Springer, 2014, pp. 740--755.

\bibitem{krishna2017visual}
R.~Krishna, Y.~Zhu, O.~Groth, J.~Johnson, K.~Hata, J.~Kravitz, S.~Chen,
  Y.~Kalantidis, L.-J. Li, D.~A. Shamma \emph{et~al.}, ``Visual genome:
  Connecting language and vision using crowdsourced dense image annotations,''
  \emph{International journal of computer vision}, vol. 123, no.~1, pp. 32--73,
  2017.

\bibitem{ordonez2011im2text}
V.~Ordonez, G.~Kulkarni, and T.~Berg, ``Im2text: Describing images using 1
  million captioned photographs,'' \emph{Advances in neural information
  processing systems}, vol.~24, 2011.

\bibitem{papineni2002bleu}
K.~Papineni, S.~Roukos, T.~Ward, and W.-J. Zhu, ``Bleu: a method for automatic
  evaluation of machine translation,'' in \emph{Proceedings of the 40th annual
  meeting of the Association for Computational Linguistics}, 2002, pp.
  311--318.

\bibitem{banerjee2005meteor}
S.~Banerjee and A.~Lavie, ``Meteor: An automatic metric for mt evaluation with
  improved correlation with human judgments,'' in \emph{Proceedings of the acl
  workshop on intrinsic and extrinsic evaluation measures for machine
  translation and/or summarization}, 2005, pp. 65--72.

\bibitem{banerjee2005automatic}
S.~Banerjee, A.~Lavie \emph{et~al.}, ``An automatic metric for mt evaluation
  with improved correlation with human judgments,'' in \emph{Proceedings of the
  ACL-2005 Workshop on Intrinsic and Extrinsic Evaluation Measures for MT
  and/or Summarization}, 2005, pp. 65--72.

\bibitem{vedantam2015cider}
R.~Vedantam, C.~Lawrence~Zitnick, and D.~Parikh, ``Cider: Consensus-based image
  description evaluation,'' in \emph{Proceedings of the IEEE conference on
  computer vision and pattern recognition}, 2015, pp. 4566--4575.

\bibitem{anderson2016spice}
P.~Anderson, B.~Fernando, M.~Johnson, and S.~Gould, ``Spice: Semantic
  propositional image caption evaluation,'' in \emph{European conference on
  computer vision}.\hskip 1em plus 0.5em minus 0.4em\relax Springer, 2016, pp.
  382--398.

\bibitem{xu2017video}
D.~Xu, Z.~Zhao, J.~Xiao, F.~Wu, H.~Zhang, X.~He, and Y.~Zhuang, ``Video
  question answering via gradually refined attention over appearance and
  motion,'' in \emph{Proceedings of the 25th ACM international conference on
  Multimedia}, 2017, pp. 1645--1653.

\bibitem{yu2019activitynet}
Z.~Yu, D.~Xu, J.~Yu, T.~Yu, Z.~Zhao, Y.~Zhuang, and D.~Tao, ``Activitynet-qa: A
  dataset for understanding complex web videos via question answering,'' in
  \emph{Proceedings of the AAAI Conference on Artificial Intelligence},
  vol.~33, no.~01, 2019, pp. 9127--9134.

\bibitem{jang2017tgif}
Y.~Jang, Y.~Song, Y.~Yu, Y.~Kim, and G.~Kim, ``Tgif-qa: Toward spatio-temporal
  reasoning in visual question answering,'' in \emph{Proceedings of the IEEE
  conference on computer vision and pattern recognition}, 2017, pp. 2758--2766.

\bibitem{goyal2017making}
Y.~Goyal, T.~Khot, D.~Summers-Stay, D.~Batra, and D.~Parikh, ``Making the v in
  vqa matter: Elevating the role of image understanding in visual question
  answering,'' in \emph{Proceedings of the IEEE conference on computer vision
  and pattern recognition}, 2017, pp. 6904--6913.

\bibitem{rennie2017self}
S.~J. Rennie, E.~Marcheret, Y.~Mroueh, J.~Ross, and V.~Goel, ``Self-critical
  sequence training for image captioning,'' in \emph{Proceedings of the IEEE
  conference on computer vision and pattern recognition}, 2017, pp. 7008--7024.

\bibitem{zhang2020object}
Z.~Zhang, Y.~Shi, C.~Yuan, B.~Li, P.~Wang, W.~Hu, and Z.-J. Zha, ``Object
  relational graph with teacher-recommended learning for video captioning,'' in
  \emph{Proceedings of the IEEE/CVF conference on computer vision and pattern
  recognition}, 2020, pp. 13\,278--13\,288.

\bibitem{zhang2021open}
Z.~Zhang, Z.~Qi, C.~Yuan, Y.~Shan, B.~Li, Y.~Deng, and W.~Hu, ``Open-book video
  captioning with retrieve-copy-generate network,'' in \emph{Proceedings of the
  IEEE/CVF Conference on Computer Vision and Pattern Recognition}, 2021, pp.
  9837--9846.

\bibitem{lin2022swinbert}
K.~Lin, L.~Li, C.-C. Lin, F.~Ahmed, Z.~Gan, Z.~Liu, Y.~Lu, and L.~Wang,
  ``Swinbert: End-to-end transformers with sparse attention for video
  captioning,'' in \emph{Proceedings of the IEEE/CVF Conference on Computer
  Vision and Pattern Recognition}, 2022, pp. 17\,949--17\,958.

\bibitem{jiang2020divide}
J.~Jiang, Z.~Chen, H.~Lin, X.~Zhao, and Y.~Gao, ``Divide and conquer:
  Question-guided spatio-temporal contextual attention for video question
  answering,'' in \emph{Proceedings of the AAAI Conference on Artificial
  Intelligence}, vol.~34, no.~07, 2020, pp. 11\,101--11\,108.

\bibitem{huang2022clover}
J.~Huang, Y.~Li, J.~Feng, X.~Sun, and R.~Ji, ``Clover: Towards a unified
  video-language alignment and fusion model,'' \emph{arXiv preprint
  arXiv:2207.07885}, 2022.

\bibitem{yang2021just}
A.~Yang, A.~Miech, J.~Sivic, I.~Laptev, and C.~Schmid, ``Just ask: Learning to
  answer questions from millions of narrated videos,'' in \emph{Proceedings of
  the IEEE/CVF International Conference on Computer Vision}, 2021, pp.
  1686--1697.

\bibitem{zellers2021merlot}
R.~Zellers, X.~Lu, J.~Hessel, Y.~Yu, J.~S. Park, J.~Cao, A.~Farhadi, and
  Y.~Choi, ``Merlot: Multimodal neural script knowledge models,''
  \emph{Advances in Neural Information Processing Systems}, vol.~34, pp.
  23\,634--23\,651, 2021.

\bibitem{yang2022zero}
A.~Yang, A.~Miech, J.~Sivic, I.~Laptev, and C.~Schmid, ``Zero-shot video
  question answering via frozen bidirectional language models,'' \emph{arXiv
  preprint arXiv:2206.08155}, 2022.

\bibitem{yan2022video}
S.~Yan, T.~Zhu, Z.~Wang, Y.~Cao, M.~Zhang, S.~Ghosh, Y.~Wu, and J.~Yu,
  ``Video-text modeling with zero-shot transfer from contrastive captioners,''
  \emph{arXiv preprint arXiv:2212.04979}, 2022.

\bibitem{oncescu2021audio}
A.-M. Oncescu, A.~Koepke, J.~F. Henriques, Z.~Akata, and S.~Albanie, ``Audio
  retrieval with natural language queries,'' \emph{arXiv preprint
  arXiv:2105.02192}, 2021.

\bibitem{xu2020crnn}
X.~Xu, H.~Dinkel, M.~Wu, and K.~Yu, ``A crnn-gru based reinforcement learning
  approach to audio captioning.'' in \emph{DCASE}, 2020, pp. 225--229.

\bibitem{chen2020audio}
K.~Chen, Y.~Wu, Z.~Wang, X.~Zhang, F.~Nian, S.~Li, and X.~Shao, ``Audio
  captioning based on transformer and pre-trained cnn.'' in \emph{DCASE}, 2020,
  pp. 21--25.

\bibitem{xu2021investigating}
X.~Xu, H.~Dinkel, M.~Wu, Z.~Xie, and K.~Yu, ``Investigating local and global
  information for automated audio captioning with transfer learning,'' in
  \emph{ICASSP 2021-2021 IEEE International Conference on Acoustics, Speech and
  Signal Processing (ICASSP)}.\hskip 1em plus 0.5em minus 0.4em\relax IEEE,
  2021, pp. 905--909.

\bibitem{koh2022automated}
A.~Koh, X.~Fuzhao, and C.~E. Siong, ``Automated audio captioning using transfer
  learning and reconstruction latent space similarity regularization,'' in
  \emph{ICASSP 2022-2022 IEEE International Conference on Acoustics, Speech and
  Signal Processing (ICASSP)}.\hskip 1em plus 0.5em minus 0.4em\relax IEEE,
  2022, pp. 7722--7726.

\bibitem{mei2021audio}
X.~Mei, X.~Liu, Q.~Huang, M.~D. Plumbley, and W.~Wang, ``Audio captioning
  transformer,'' \emph{arXiv preprint arXiv:2107.09817}, 2021.

\bibitem{liu2022leveraging}
X.~Liu, X.~Mei, Q.~Huang, J.~Sun, J.~Zhao, H.~Liu, M.~D. Plumbley, V.~Kilic,
  and W.~Wang, ``Leveraging pre-trained bert for audio captioning,'' in
  \emph{2022 30th European Signal Processing Conference (EUSIPCO)}.\hskip 1em
  plus 0.5em minus 0.4em\relax IEEE, 2022, pp. 1145--1149.

\bibitem{ding2022davit}
M.~Ding, B.~Xiao, N.~Codella, P.~Luo, J.~Wang, and L.~Yuan, ``Davit: Dual
  attention vision transformers,'' \emph{arXiv preprint arXiv:2204.03645},
  2022.

\bibitem{he2020deberta}
P.~He, X.~Liu, J.~Gao, and W.~Chen, ``Deberta: Decoding-enhanced bert with
  disentangled attention,'' \emph{arXiv preprint arXiv:2006.03654}, 2020.

\bibitem{li2020oscar}
X.~Li, X.~Yin, C.~Li, P.~Zhang, X.~Hu, L.~Zhang, L.~Wang, H.~Hu, L.~Dong,
  F.~Wei \emph{et~al.}, ``Oscar: Object-semantics aligned pre-training for
  vision-language tasks,'' in \emph{European Conference on Computer
  Vision}.\hskip 1em plus 0.5em minus 0.4em\relax Springer, 2020, pp. 121--137.

\bibitem{wang2021ufo}
J.~Wang, X.~Hu, Z.~Gan, Z.~Yang, X.~Dai, Z.~Liu, Y.~Lu, and L.~Wang, ``Ufo: A
  unified transformer for vision-language representation learning,''
  \emph{arXiv preprint arXiv:2111.10023}, 2021.

\bibitem{zhang2021vinvl}
P.~Zhang, X.~Li, X.~Hu, J.~Yang, L.~Zhang, L.~Wang, Y.~Choi, and J.~Gao,
  ``Vinvl: Revisiting visual representations in vision-language models,'' in
  \emph{Proceedings of the IEEE/CVF Conference on Computer Vision and Pattern
  Recognition}, 2021, pp. 5579--5588.

\bibitem{dou2022empirical}
Z.-Y. Dou, Y.~Xu, Z.~Gan, J.~Wang, S.~Wang, L.~Wang, C.~Zhu, P.~Zhang, L.~Yuan,
  N.~Peng \emph{et~al.}, ``An empirical study of training end-to-end
  vision-and-language transformers,'' in \emph{Proceedings of the IEEE/CVF
  Conference on Computer Vision and Pattern Recognition}, 2022, pp.
  18\,166--18\,176.

\bibitem{yao2021filip}
L.~Yao, R.~Huang, L.~Hou, G.~Lu, M.~Niu, H.~Xu, X.~Liang, Z.~Li, X.~Jiang, and
  C.~Xu, ``Filip: Fine-grained interactive language-image pre-training,''
  \emph{arXiv preprint arXiv:2111.07783}, 2021.

\bibitem{li2022blip}
J.~Li, D.~Li, C.~Xiong, and S.~Hoi, ``Blip: Bootstrapping language-image
  pre-training for unified vision-language understanding and generation,''
  \emph{arXiv preprint arXiv:2201.12086}, 2022.

\bibitem{hu2022scaling}
X.~Hu, Z.~Gan, J.~Wang, Z.~Yang, Z.~Liu, Y.~Lu, and L.~Wang, ``Scaling up
  vision-language pre-training for image captioning,'' in \emph{Proceedings of
  the IEEE/CVF Conference on Computer Vision and Pattern Recognition}, 2022,
  pp. 17\,980--17\,989.

\end{thebibliography}
}


%







\end{document}